\documentclass[letterpaper]{article} 
\usepackage{aaai24}  
\usepackage{times}  
\usepackage{helvet}  
\usepackage{courier}  
\usepackage[hyphens]{url}  
\usepackage{graphicx} 
\usepackage{natbib}  
\usepackage{caption} 
\usepackage{algorithm}
\usepackage{algorithmic}
\nocopyright
\usepackage{newfloat}
\usepackage{listings}
\DeclareCaptionStyle{ruled}{labelfont=normalfont,labelsep=colon,strut=off} 
\floatstyle{ruled}
\newfloat{listing}{tb}{lst}{}
\floatname{listing}{Listing}
\usepackage{multirow}
\usepackage{tikz}
\usepackage{xcolor}
\usepackage{booktabs, multirow, tabularx, threeparttable}
\usepackage{subcaption}

\usepackage{amsmath}
\usepackage{enumitem}
\usepackage{cancel}
\usepackage{soul}
\usepackage{utfsym}
\usepackage{pifont}
\newcommand{\cmark}{\ding{51}}%
\newcommand{\xmark}{\ding{55}}%

\title{Hidden or Inferred: Fair Learning-To-Rank with Unknown Demographics}

\author {
    Oluseun Olulana\textsuperscript{\rm 1}, Kathleen Cachel\textsuperscript{\rm 1}, Fabricio Murai\textsuperscript{\rm 1,\rm 2}, Elke Rundensteiner\textsuperscript{\rm 1,\rm 2}}

\affiliations {
    \textsuperscript{\rm 1}Data Science Program, Worcester Polytechnic Institute, Worcester, MA,\\
    \textsuperscript{\rm 2}Computer Science Department, Worcester Polytechnic Institute, Worcester, MA\\
    \{omolulana, kcachel, fmurai, rundenst\}@wpi.edu
}

\begin{document}

\newcommand{\crossmark}{\scalebox{0.75}{\usym{2613}}}

\newcommand{\adv}{$g_{adv}$}
\newcommand{\dadv}{$g_{dis}$}
\newcommand{\smone}{\dadv $\leftrightarrow$ \adv}
\newcommand{\smtwo}{\dadv $\rightarrow$ \adv}
\newcommand{\smthree}{\dadv $\leftarrow$ \adv}
\newcommand{\dadexp}{Dadv/Adv exposure ratio}

\newcommand{\algoone}{\textsc{Oblivious}}
\newcommand{\algotwo}{\textsc{LTR}}
\newcommand{\algothree}{\textsc{Hidden}}
\newcommand{\algofour}{\textsc{FairLTR}}
\newcommand{\algofive}{\textsc{Oblivious+FairRR}}
\newcommand{\algosix}{\textsc{LTR+FairRR}}
\newcommand{\algoseven}{\textsc{Hidden+FairRR}}

\newcommand{\ds}{Data Science Program, Worcester Polytechnic Institute, Worcester, MA (\{omolulana, kcachel\}@wpi.edu).}

\maketitle

\begin{abstract}
As learning-to-rank models are increasingly deployed for decision-making in areas with profound life implications, 
the FairML community has been developing fair learning-to-rank (LTR) models. These models rely on the availability of sensitive demographic features such as race or sex. However, in practice, regulatory obstacles and privacy concerns protect this data from collection and use.
As a result, practitioners 
may either need to promote fairness despite the absence of these features or
turn to demographic inference tools to attempt to \textit{infer} them. 
Given that these tools are fallible, this paper aims to further understand how errors in demographic inference impact the fairness performance of popular fair LTR strategies.
In which cases would it be better to keep such demographic attributes \textit{hidden} from models versus \textit{infer} them?
We examine a spectrum of fair LTR strategies
ranging
from fair LTR  with and without demographic features hidden  versus inferred
to  
fairness-unaware LTR 
followed by fair re-ranking.
We conduct a controlled empirical investigation 
modeling different levels of inference errors
by systematically perturbing the inferred sensitive attribute.  
We also perform three case studies with real-world datasets and popular 
open-source inference methods. 
Our findings  reveal that as inference noise grows,  LTR-based methods that incorporate fairness considerations into the learning process may  increase bias.
In contrast,  fair re-ranking strategies are more robust to inference errors. 
All source code, data, and experimental artifacts 
of our experimental study 
are available here: {\it{https://github.com/sewen007/hoiltr.git}}
\end{abstract}

\section{Introduction}
\textbf{Background: Fairness of LTR-based Ranking.}
Ranked search results are increasingly at the heart of artificial intelligence and automated decision-making systems. As such systems progressively impact our daily lives, there is a growing need to ensure that these technologies do not disproportionately harm or replicate societal biases toward disadvantaged populations or legally protected groups. To this end, the 
fair machine learning
community has developed various fair LTR models \cite{DELTR,wang2022meta} and metrics \cite{fairpatro22, ekstrand2021fairness} for assessing 
such models. At a high level, LTR models learn a scoring function, so that when deployed, the model's learned relevance scores produce an ordering of candidate items. While conventional fairness-unaware LTR methods aim to produce a utility maximizing ordering of candidates, 
{\it fairness-aware LTR} methods aim to ensure this ranking is also a {\it fair} ordering of candidates (items).

\textbf{Challenges: Fairness without 
Protected Attributes.}
Even with such progress,
practical obstacles prevent the widespread adoption of these bias mitigation methods. One  challenge is that fair LTR models as well as other fairness-enhanced methods, such as fair classifiers, require demographic information associated with candidate items during both model training and subsequent
model deployment for real-world use \cite{dwork2012fairness}. However, in practice, it may not be possible to collect, gain access, or use protected demographic features due to privacy concerns or legal restrictions. For instance, the European Union's GDPR (General Data Protection Regulation) legislation strictly regulates collection, retention, and use of demographic data for algorithmic purposes \cite{EUdataregulations2016}.

Also, there is policy tension between mandates {\it to ensure algorithms are fair} and mandates {\it prohibiting the use of demographic data}.
One example is the United States credit industry \cite{ecfr}. 
Consumer-facing lenders are largely prohibited from collecting 
sensitive demographic data; yet United States federal law explicitly prohibits creditors from discriminating on the basis of certain protected demographics
and thus leaves them wanting to verify that they did not do so -- the later of which requires access to the same sensitive information \cite{bogen2020awareness}. Thus, companies and institutions are increasingly caught in the middle. Surveys of data scientists and algorithm developers highlight the challenge these practitioners face in promoting fairness without such sensitive data \cite{Holstein2018ImprovingFI}. 

\textbf{State-of-Art and their Limitations.}
A common workaround 
is to 
infer demographic data from other available information, such as first names, social media  \cite{socialmediainferring}, or  email content \cite{emailgender}. 
However, the accuracy of these estimation (inference) tools can vary significantly
\cite{emailgender, accuracycomparison}. 

Recent work in the fair-ranking community has developed fair ranking metrics \cite{ghazimatin2022measuring, kirnap2021estimation} that account for error rates of demographic attributes. \citet{ghosh2021fair} investigate the Deterministic Constrained Sorting (DetConstSort) re-ranking method coupled with the use of demographic inference for integrating fairness. DetConstSort is a fair re-ranking algorithm used as  post-processing step.
Their study reveals that DetConstSort performs poorly when given inaccurate demographic information.  This  raises questions regarding the performance of fair LTR strategies that instead choose to either infer or ignore the demographic information. Fair LTR methods have been shown to achieve better fairness-relevance trade-offs compared to applying fair re-ranking methods like DetConstSort to existing fairness-unaware LTR methods -- while assuming full access to 
 protected (demographic) attributes \cite{DELTR}.  This makes these approaches  desirable in real-world settings. 
Therefore, with significance to practitioners, in this work, we ask:
{\it How do 
errors in demographic inference impact the fairness performance of different fair LTR strategies? Also, how 
do these errors impact utility?}

\textbf{Our Approach.}
We tackle these research questions by investigating the performance of popular fair LTR strategies 
when deployed in situations where demographic data is unavailable and thus
needs to be inferred.
We investigate both fair LTR and fair re-ranking type models, as they (and their combinations) cover the majority of fairness-enhanced rank-based machine learning pipelines.
We also investigate the case where 
protected attributes remain hidden as a fairness strategy, meaning, all candidates have the
``unknown'' as demographic attribute.
We study
real-world scenarios where models are first trained on available data, with and without sensitive demographic attributes. 
However, later, during model deployment, data issues may arise that saddle practitioners with a choice to make -- to work with data while 
the sensitive attribute remains \textit{hidden} or to augment the data by \textit{inferring} the missing demographic attributes. While our experiments focus on gender, our study methodology is equally applicable to  other protected features.

Our  experimental evaluation 
is composed
of two  studies (Sections \ref{simulations} and \ref{cases}).
In the first study, we systematically perturb the inferred protected attribute to model different levels of inference error under three scenarios. Each scenario is designed
to capture the different kinds and levels of errors possible in a real-world scenario.
In the second study, we make use of popular demographic inference services \cite{genderapi, nmsor, btn} 
to compare and contrast their impact
in the context of 
three  real-world data sets. 
We explore
the effect on fairness with respect to ranked candidates' true group identities. 

Our investigation has led to the following findings:
\begin{itemize}
    \item
    Re-ranking fair strategies that enforce group fairness based on the inferred distribution of test candidates are more robust to inaccurate inference of demographic attributes than fair LTR strategies. This leads to the guidance that, under noisy inference scenarios, practitioners may  achieve a higher level of fairness if adopting a fair re-ranking  instead of a fair LTR strategy.

    \item
    While fairness-aware strategies achieve considerable fairness even when working with inaccurately inferred demographic attributes, fairness decreases as inference errors increase. This suggests that a practitioner's fairness goals may be adversely affected by the utilization of lower-quality inference tools. 
    Practitioners are  urged to proceed with caution and verify  the quality of their inference tools beforehand.
    
    \item In a scenario where demographic attributes are missing,
it is better
to utilize a fairness-aware model that relies on inferred missing attributes than to adopt a fairness-unaware model that ignores the missing attributes as long
    as  inference errors are relatively low (up to $10\%$).
    
    \item 
   Lastly, we 
   observe that across all  three real-world data set case studies, 
   the  fairness-unaware models
   show increased levels of fairness when working with demographic attributes that had been incorrectly inferred.
   We attribute this phenomenon to
   candidate items
   being wrongly classified as members of the alternate group.
\end{itemize}

\section{Related Work}

While  many ethical considerations in designing ranking systems exist, fairness is typically conceptualized as either individual fairness or group fairness. Individual fairness ensures that similar individuals receive similar outcomes \cite{dwork2012fairness}. Group fairness ensures that protected groups of people (such as race or gender) receive comparable shares of the positive outcome 
\cite{li2021user, ekstrand2021fairness}. In this work, we consider group fairness, the primary concern of the fair ranking literature \cite{fairpatro22},
which aligns with the existing focus of AI regulation \cite{jillson2021aiming}. 

Fairness can be incorporated into an LTR model by adding a fairness  parameter or constraint to the learning objective \cite{DELTR, singh2019policy, wang2022meta}. Alternatively, instead of modifying a fairness-unaware LTR algorithm, fairness can be introduced by reordering a ranking generated by a model \cite{geyik, zehlike2017fa}. When striving to be demographically fair, not only is the fairness affected, but also the utility (relevance) of the ranking. In other words, integrating fairness into a ranking framework induces a  fairness-utility trade-off. This trade-off has been addressed in some existing work on fairness 
\cite{DELTR, li2022fairness}. \citet{DELTR}, in their fairness-aware LTR loss function, introduce a parameter $\gamma$ that balances the trade-off between utility and disparate exposure.
\cite{li2022fairness} emphasize the importance for researchers to explore the relationship between fairness and utility to motivate practitioners to promote fairness. In an earlier paper by \cite{li2021user}, a constrained optimization problem was formulated, where the overall recommendation quality (utility) is the objective function and an upper bound $\epsilon$ on group (un-)fairness is enforced via a constraint.
The overarching assumption underlying fairness-enhanced algorithms is that demographic data is readily accessible 
and correct for use by these algorithms \cite{Holstein2018ImprovingFI}. 
Recent work has begun to relax this assumption by developing methods that account for
error rates of demographic attributes when incorporating fairness~\cite{mehrotra2022fair, celis2021fair, wang2020robust, mozannar2020fair} and 
by designing algorithms 
that rely on latent feature representations instead of explicit demographic information \cite{pmlr-v80-hashimoto18a, Lahoti2020FairnessWD}.
\citet{zhang2021assessing} focus on fairness with
incomplete data in classification and regression tasks. Their analysis relies on 
subselecting only data points (rows) where none of its values are missing. To analyze the importance of factoring missing data into the classifier models, \citet{goel2021importance} studied fairness guarantees in the training procedures under various distributions. The study showed that incorporating data ``missingness'' can help determine the choice of fairness design principles to use in practice. To tackle information inefficiency, \citet{noriega2019active} proposed 
to acquire information based on the need of the group in fair classification. They showed that this helped achieve major fairness objectives, for example, equal opportunity. However, these and other algorithms as well as their empirical evaluations predominantly focus on either {\it fair classification}  \cite{pmlr-v80-hashimoto18a, celis2021fair, wang2020robust, mozannar2020fair, ghosh2023fair} or are restricted to re-ranking (i.e., re-ranking an existing ranking) \cite{ghosh2021fair, mehrotra2022fair}.  While \citet{ghosh2021fair} explore 
dealing with uncertainty in fair ranking algorithms and, in a later study, for fair classification \cite{ghosh2023fair},
the relative performance of alternate fair LTR strategies in the presence of unknown and inferred demographic groups
remains an open question.

\section{Experimental Methodology}
\label{sec:methodology}
We introduce fair ranking algorithms, and then describe
how we compose these algorithms into a spectrum of alternate strategies for integrating fairness into fair-learning-to-rank pipelines. Next, we  present tools  for inferring protected attributes, followed by metrics for fairness and utility.

\subsection{Preliminaries}
 To train an ranking model, we start with a list $C = \{c_1, ..., c_n\}$ where each candidate item $c_i$ is associated with an attribute score vector $x_i$ and a ground-truth relevance score $s(c_i)$ (a.k.a., judgment score).
These and other useful notation is presented in Table \ref{tab:notation}.
\begin{table}[ht]
    \centering
    \begin{tabular}{cl}
        \toprule
         Symbol & Definition \\
        \midrule
        $C$& List of items $\{c_1, ..., c_n\}$ to be ranked\\
        $x_i$ & Attribute score vector for $c_i$ \\
        $\tau$ & A ranking ordering (top is better)\\
        $s(c_i)$  & Ground-truth score of item  $c_i \in C$ \\
        $s_{\tau}(j)$  & Judgment score of item at position $j \in \tau$ \\
        $g_{dis}$ & Disadvantaged group \\
$g_{adv}$ & Advantaged group\\
        \bottomrule
    \end{tabular}
    \caption{Notation table}
    \label{tab:notation}
\end{table}

\subsection{State-Of-The-Art Fairness Interventions}\label{sec:model_desc}
In this section, we introduce the specific instantiations
of the fair LTR models and the fair re-ranking models that we study, namely, DELTR (Disparate Exposure In Learning-to-Rank)

and  DetConstSort \cite{geyik},
respectively. We however, begin by introducing a fairness-unaware model, Listnet \cite{cao2007learning}. 
\begin{description}[wide,labelindent=0pt,parsep=0pt,itemsep=5pt,style=unboxed]
\item[Listnet] Proposed by \citet {cao2007learning}, it defines a loss function based on the
``top one probability'', defined as the probability for an item to be ranked at the top given the scores of all items.
Given a list $C$ with corresponding $x_i$ and  $s(c_i)$ values, the model is trained to assign judgment scores to unseen candidate items with attribute scores. The judgment scores can then be used to rank the items in relative order of relevance.

\item[DELTR] Proposed by \citet{DELTR}, we choose it to represent fairness-aware LTR models.
This method aims to reduce disparate exposure (a measure of unfairness), while simultaneously reducing rank prediction errors. Conceptually, given a list $C$ with corresponding $x_i$ and  $s(c_i)$ values, it is assumed that each candidate belongs to one of two disjoint groups, one of which is \textit{protected} ($g_{dis}$). 
A group with higher visibility at the top of the ranking than another is said to have a higher exposure.
This method also assumes that disparate exposure is experienced by $g_{dis}$.
The model is trained to reduce unfairness, while aiming to maintain accurate score predictions.
During training, DELTR learns to assign judgment scores to candidates using the candidates's attributes (including their protected attributes, e.g., sex) that are provided to the model.  The trained model can then assign new judgment scores to unseen candidate items based on their corresponding attributes.

\item[DetConstSort] A fair re-ranking algorithm that works to improve fairness by enforcing group representation within the top $k$ positions of a ranking. Given a list of candidates ranked by their predicted scores and a list of groups $\mathcal{G}$, DetConstSort re-ranks the list such that, for all groups $g \in \mathcal{G}$ and for all $k$ representing 
a position on the ranking, the number of candidates in group $g$ among the top $k$ results is at least $\lfloor p_{g}\times k\rfloor$, where $p_g$ is a target proportion of candidates from group $g$.  Most commonly, the target population corresponds to the underlying distribution $P_C$ of the candidate set. Unlike DELTR, DetConstSort is a deterministic algorithm that does not require training. 
\end{description}

\subsection{Spectrum of Fair LTR Strategies}
\label{sec:approaches_desc}

Next, we describe the spectrum of alternate strategies we study
for integrating fairness into popular LTR algorithms when the protected attribute is not known at test time.

\subsubsection{Fairness Strategies}

\label{sec:approaches}
Table \ref{tab:overview_strategies} presents comprehensive details regarding the training, testing (ranking), and re-ranking aspects associated with each strategy.
For each strategy, the model \emph{training} may or may not include a protected attribute. 
For models trained with the protected attribute, we assume
that the protected attributes
are  not available \emph{during testing}. This leaves two possibilities for imputation during ranking and re-ranking: 
either (i)  \emph{inferring} the protected attribute or (ii)  \emph{hiding} it which means applying the model
without gaining access to the value. 

\begin{table*}
  \centering
   
 \begin{tabular}{llccc}
\toprule
\multirow{3}{*}{Fairness}&\multirow{3}{*}{{Strategies}} &\multicolumn{3}{c}{Protected Attribute Use}\\ 
&&Training & Testing & Re-ranking\\
\midrule
\multirow{3}{*}{Unaware}  & \textbf{\algoone{}}& n/a & n/a & n/a  \\
&\textbf{\algotwo{}}& ground truth & inferred & n/a \\
&\textbf{\algothree{}}&ground truth & hidden & n/a \\
\midrule
\multirow{4}{*}{Aware}&\textbf{\algofour{}}&ground truth & inferred & n/a \\
&\textbf{\algofive{}}& n/a & n/a & inferred \\
&\textbf{\algosix{}}&ground truth & inferred & inferred (same) \\

&\textbf{\algoseven{}}&ground truth & hidden & inferred \\
\bottomrule 
\end{tabular}%

\caption{Fair ranking strategies and how they use the protected attribute  \textbf{during training, testing (ranking) and re-ranking}: inferred (via noise model or inference tool), hidden (attribute replaced by constant value for all candidates), n/a.}
\label{tab:overview_strategies}
\end{table*}
We consider three possible ways of generating the input ranking provided to the fair re-ranking algorithm, DetConstSort:  
(i) using a ListNet model trained without the protected attribute; or (ii) training a ListNet model using ground-truth protected attribute values and, during testing, either inferring the protected attribute or (iii) hiding it by replacing it with a constant value for all test candidates.
  For comparison, we also consider the three cases when the rankings were generated without the re-ranking performed by DetConstSort. These variants and DELTR altogether sum up to a total of seven strategies. We describe them in detail next;
while introducing their acronyms seen in Table~\ref{tab:overview_strategies}.

\begin{description}[wide,labelindent=0pt,parsep=0pt,itemsep=5pt,style=unboxed]
\item[\algoone{} \label{algo1} ]   This baseline stands for the approach where, \textit{during both training and testing,  a fairness-unaware ranking model has no access to the protected attributes}. The model is thus said to be ``oblivious'' to the protected attributes. In this setting, we use the fairness-unware  
Listnet.
\item[\algotwo{} \label{algo2}] 
In this approach, a \textit{fairness-unaware model is trained with access to the ground-truth protected attribute}, and \textit{during testing it relies on inferred protected attributes}. ListNet is also used for this model. Note that the underlying ranking model Listnet does not consider fairness. Hence, we utilize this approach in isolation as another 
 fairness-unaware baseline (different from \algoone{}).

\item[\algothree{}] \label{algo3} 
This strategy is to train \textit{a fairness-unaware model  with access to the ground-truth protected attributes}, and \textit{to hide the protected attributes during testing}. For this model we also use Listnet. This approach differs from \algotwo{} because instead of inferring the protected attribute, \algothree{} neutralizes any impact of the attribute's group value on the ranking decision by replacing it by the same constant value for all candidates.
This causes the model to ignore the direct effect of the protected attribute value. This approach does not rely on actual or inferred protected attribute values and is thus invariant to inference errors.

\item[\algofour{}]

\label{deltr-intro} 
In this approach, \textit{a fairness-aware model is trained 
with ground-truth protected attributes} and \textit{during testing, the protected attributes are inferred}.
For this we use the DELTR model.
\item[\algofive{}]
This  stands for the approach where a \textit{fairness-unaware ranking model trained without the protected attribute} is used to rank the list.  
Then \textit{this list is processed by a fair re-ranking algorithm that uses an inferred protected attribute}.
We use Listnet as the fairness-unaware ranking model and DetConstSort as the fair re-ranking algorithm. As stated in 
Section \ref{sec:approaches}, DetConstSort relies on group proportions to ensure fairness, for this we use the inferred group proportions. 
\item[\algosix{}] 
In this approach, \textit{a fairness-unaware ranking model} (trained with ground-truth protected attributes, but using inferred attribute during testing) is  \textit{followed by a fair re-ranking algorithm which also works with the same inferred protected attributes}. We utilize ListNet for the first step, and DetConstSort as the second. As above, DetConstSort uses the inferred group proportions. This approach is similar to \algofive{}, however the fairness-unaware model used in the first stage is trained in the presence of the protected attributes.

\item[\algoseven{}]\label{algo7}
This approach utilizes \textit{a fairness-unaware ranking model} (trained with  protected attributes that are then hideen during testing) followed by a \textit{fair re-ranking algorithm which also works with the same inferred protected attributes}. We also utilize ListNet for the first step, and DetConstSort as the seond step. As in \algofive{} and \algosix{}, DetConstSort uses the inferred group proportions. 
\end{description}

\subsection{Inference  for Missing Protected Attributes}\label{sec:services}

Inference services use other available
attributes of the candidates,   such as names, email, or images, to infer demographic attributes such as sex, age, or race of a person. While many such services exist, we characterize  three popular 
solutions  for inferring sex used in our experiments. 
In Section \ref{cases},
 we evaluate their  accuracy on the datasets studied  in this paper.

\begin{description}[wide,labelindent=0pt,parsep=0pt,itemsep=5pt]
\item[Behind The Name] 
This service\footnote{\url{https://www.behindthename.com/}} gives the user the option to use only the first name or the full name to deduce sex. It utilizes the etymology (meaning) of a name and history of names to infer sex. It covers various regions of the world with names collated from national statistics agencies.

\item[Gender API] 
This service\footnote{\url{https://gender-api.com/}} uses either the first name or a combination of both first and last names to deduce sex. 
The system can also leverage supplementary parameters, such as
an email address or location (country, IP address, and browser) along with publicly accessible governmental data, and social media data.

\item[Namsor]
This service\footnote{\url{https://namsor.app/}} uses both the first and last name to infer sex. It claims a broad coverage across all languages, alphabets, countries, and regions. The foundational data is sourced from a compilation of 1.3 million  names extracted from baby name statistics, encompassing various countries, morphology, languages, and ethnicity.
\end{description}

\textbf{Names not Recognized By  Inference Service}. Each inference service returns 
candidates whose protected attribute could not be determined with a degree of certainty above a threshold,
here called unknowns.
Since these attributes are required by the fairness-aware strategies,
 we will need to 
assign some  valid  value
to these {\it unknowns} (see Section~\ref{cases}).
\subsection{Metrics of Fairness and Utility}\label{sec:metrics}
We employ the following metrics in our analysis, encompassing both established fairness and utility measures. 

\begin{description}[wide,labelindent=0pt,parsep=0pt,itemsep=5pt]

\item[Rank Fairness Metrics]  
We select three primary metrics to assess rank fairness, two from a representation  (Skew and NDKL) and one from an exposure standpoint. 

\begin{itemize}
    \item {\it Skew}~\cite{geyik}.
    The skew is a fairness metric used for determining the (dis)advantaged group. We assume that candidates benefit more from being at the top of the ranking. Skew is defined for a group $g \in \mathcal{G}$ in ranking $\tau$ measured at position $k$ as:
\begin{equation}
\label{eqn:skew}
    Skew_g(\tau)@k = \frac{p_{\tau@k,g}}{p_{C,g}},
\end{equation}
where $p_{\tau@k,g}$ is the proportion of items belonging to $g$ within the top $k$ items in ranking $\tau$, and $p_{C,g}$ the proportion of items belonging to $g$ in the entire ranked candidate set $C$. A skew value of $1$ is best and indicates that the group $g$'s proportion of the top $k$ positions in $\tau$ is the same as its proportion in the item set $C$. 
Values above~1 indicate group $g$ is over-represented and values below~1 indicate group $g$ is under-represented.
\item 
{\it NDKL}~\cite{geyik}\label{sec:ndkl}.
The Normalized Discounted Kullback-Leibler
divergence of a ranking $\tau$ given a set of groups $\mathcal{G}$ is: 
\begin{equation}\label{eqn_ndkl}
    \mathit{NDKL}@k(\tau)= \frac{1}{Z}\sum^{k}_{i = 1}\frac{1}{\log_{2}(i +1 )}d_\mathit{KL}(P_{\tau@k} || P_{C}),
\end{equation}
where $d_\mathit{KL}(P_{\tau@k} || P_{C})$ is the KL-divergence between $P_{\tau@k}$, the (discrete) distribution of group proportions in the first $k$ positions of $\tau$ and $ P_{C}$, the distribution of group proportions in the entire item set $C$. Then 
$Z = \sum_{i = 1}^{k} \frac{1}{\log_2(i + 1)}$. NDKL ranges from $0$ to $\infty$, where lower value of zero indicates that, at all prefixes of ranking $\tau$, all groups are represented proportionally.
 NDKL assesses fairness across all groups and does not indicate which group is over- or under-advantaged.

\item 
{\it Average Exposure} \cite{singh2018fairness} and  Exposure Ratio~\cite{DELTR}.
The DAdv/Adv Exposure Ratio of a group $g_{dis}$ relative to another group $g_{adv}$ in ranking $\tau$ is:
\begin{equation}\label{eqn_ave_exp_r}
    \mathit{DAdv/Adv~Exp~Ratio}(\tau) = \frac{\mathit{Avg Exposure}(\tau,g_{dis})}{\mathit{Avg Exposure} (\tau,g_{adv})},
\end{equation}
where Average Exposure for group $g$ in ranking $\tau$ is $\mathit{Avg Exposure}(\tau,g) = \sum_{c_i \in g}\mathit{Exposure}(\tau,c_i)/|g|$ and the exposure of item $c_i$ in ranking $\tau$ is $\mathit{Exposure}(\tau,c_i) = 1 / \log_2(\tau(c_i)+1))$.

The ideal DAdv/Adv Exp-Ratio is 1 indicating both groups have the same average exposure in ranking $\tau$. A ratio below 1 means group $g_{dis}$ is under-exposed in $\tau$ (i.e., unfairly disadvantaged) and a value above 1 means $g_{dis}$ is over-exposed (i.e., unfairly advantaged) in $\tau$.
\end{itemize}
\item[Utility Metric.] 
The utility of a ranking captures the relevance of the ordered items with respect to a criterion.
\begin{itemize}
   {\it
    \item NDCG} \cite{jarvelin2002cumulated}.
The Normalized Discounted Cumulative Gain of a ranking $\tau$ is given as:
\begin{equation}
    \mathit{NDCG}@k(\tau)= \frac{1}{Z}\sum_{i=1}^{k}\frac{s_{\tau}(i)}{\log_2(i+1)},
\end{equation}
where $s_\tau(i)$ is the score of the $i$-th element in the ranked list $\tau$
and 
$Z=\sum_{i=1}^k\frac{1}{\log_2(i+1)}.$  
NDCG gives a sense of the order the documents in terms of their relevance. A value of 1 denotes the ranking has the highest utility (i.e., it orders items by decreasing scores) and as NDCG decreases down to 0, $\tau$ provides less utility.
\end{itemize}
\end{description}

\section{Experimental Design}\label{sect.experiments}
\subsection{Data Sets}\label{sec:datasets}

We train our models using real-world datasets containing the ground truth for the protected attribute ``sex''. Before training, we randomly split our data into training and testing (ranking) datasets 
 using an 80/20 split, respectively. In the controlled study described in Section~\ref{simulations}, four real-world
datasets -- Law, NBA/WNBA, COMPAS and Boston Marathon -- are used in the experiments. Since the Law dataset lacked attributes (names) suitable for inferring the protected attribute, we have only used the other three data sets in the second study described in Section~\ref{cases}.

\begin{description}[wide,labelindent=0pt,parsep=0pt,itemsep=5pt]
    \item[Law] This dataset was obtained from the original DELTR \cite{DELTR} experimental repository. It was initially derived from a study conducted by \citet{wightman1998lsac} to assess potential bias against minorities in LSAT scores. It consists of anonymized information from first-year students at some law schools. 
    
    \item[COMPAS] This dataset was collected by Propublica in their analysis of COMPAS (Correctional Offender Management Profiling for Alternative Sanctions) tool \cite{ProPublica}. It contains attributes used for predicting the likelihood for a criminal defendant to offend again. 
    
    \item[(W)NBA] This dataset contains information about WNBA (Women's National Basketball Association) and NBA (National Basketball Association) players. Each player is associated with career points which determine their position in the ranking. Other attributes include number of seasons they played in the career and their average player efficiency ratio.
    \item[Boston Marathon] This dataset contains a list of Boston Marathon finishers in 2019. It contains both male and female runners. Each runner's times at 7 different stages were used in our experiment \footnote{\url{kaggle.com/datasets/daniboy370/boston-marathon-2019}}.

\end{description}

Table~\ref{tab:dataset} shows the number of males and females in each ranking dataset during deployment (testing). Each test set represents 20\% of the respective complete dataset.

\begin{table*}[ht]
\centering

\begin{tabular}{c|r|rrr|cc|cr|c}
\toprule                   \multirow{2}{*}{Dataset}
& Training & \multicolumn{3}{c|}{Test}   & \multirow{2}{*}{\dadv} & Inference  &\multicolumn{2}{c|}{Study used in}& \multirow{2}{*}{Target feature} \\ 
& \multicolumn{1}{c|}{Size} &Males &Females & Size &&attribute&Controlled &Case &  \\\midrule

\textbf{Law} & 4,882 & 55\% & 45\% & 1,221  & females & \underline{not available} &\cmark& \xmark    & first year scores \\
\textbf{(W)NBA}& 3,726 &78\%  & 22\% & 992   & females & names &\cmark&\cmark & career points \\
\textbf{COMPAS}& 4,917 &82\% &  18\% & 1,257    & males & names &\cmark&\cmark    &  recidivism score  \\
 \textbf{Boston M}& 21,063 &55\% &  45\% & 5,226    & females & names &\cmark&\cmark                     & official time  \\
\bottomrule
    \end{tabular}
\caption{Statistics of the training and test data, disavantaged group designation, attribute used for inference and set of experiments in which each data set was used.}    
    \label{tab:dataset}
\end{table*}

\begin{table*}
  \centering
 \begin{tabular}{cl|c|rrr|rrr}
\toprule
& \multirow{2}{*}{Service}  &  Test & \multicolumn{3}{c|}{Have Inference Result} & \multicolumn{3}{c}{Inference Result Not Available}  \\
& & Size & Correct & Incorrect & Total &\dadv & \adv & Total  \\\midrule
\parbox[t]{2mm}{\multirow{3}{*}{\scriptsize\rotatebox[origin=c]{90}{(W)NBA}}} & \textbf{Gender API}    & &          931 (94\%) & 39 (4\%) & 970 (98\%)&   7 (.7\%)              & 15 (2\%) & 22 (2\%)      \\
& \textbf{Behind The Name}     &  992 &     746 (75\%) & 8 (1\%) & 754 (76\%)     &     81 (8\%)                   & 157 (16\%) & 238 (24\%) \\
& \textbf{Namsor}   &  & 397 (40\%) & 0 (0\%) & 397 (40\%)         &  131 (13\%)             & 464 (47\%) & 595 (60\%) \\ \hline

\parbox[t]{2mm}{\multirow{3}{*}{\scriptsize\rotatebox[origin=c]{90}{COMPAS}}} & \textbf{Gender API}    & & 1135 (90\%) & 72 (6\%) & 1207 (96\%)           &   39 (3\%)                & 11 (1\%) & 50 (4\%)\\
& \textbf{Behind The Name}     & 1,257 & 905 (72\%) & 38 (3\%) & 943 (75\%)        & 239 (19\%) & 75 (6\%) &314 (25\%)                  \\
& \textbf{Namsor}   &  & 500 (40\%) & 15 (1\%) & 515 (41\%)         & 608 (48\%)   & 134 (11\%) & 742 (59\%)\\\hline

\parbox[t]{2mm}{\multirow{3}{*}{\scriptsize\rotatebox[origin=c]{90}{BostonM}}} & \textbf{Gender API}    & &  5019 (96\%) & 155 (3\%) & 5174  (99\%)        &  23 (.4\%)   & 29 (1\%) &52 (1\%) \\
& \textbf{Behind The Name}     & 5,226&   4302 (82\%) & 88 (2\%) & 4390 (84\%)        & 443 (8\%) & 394 (8\%)  & 836 (16\%)        \\
& \textbf{Namsor}   & &    2090 (40\%) & 0 (0\%) & 2090 (40\%)       & 1474 (28\%)   & 1662 (32\%) &3136 (60\%)  \\ 
\bottomrule
\end{tabular}

\caption{Statistics of inferred protected attribute for each data set and service. Candidate items with inferred protected attributes are grouped by inference correctness. Candidate items for failed  inference are grouped based on ground-truth attribute values.}
\label{inaccuracies}
\end{table*}

\subsection{Disadvantaged Groups in Our Datasets}
\label{skew_calc}
To identify the disadvantaged group within each of our datasets when candidates are ordered by ground-truth judgment scores, we compute the {\it skews} for each group at every position in the ranking
(Equation~\ref{eqn:skew}).
Graphs displaying the distribution of these groups throughout each of the datasets are included in the supplemental material (see Figure \ref{fig:skew_graphs}).
The group displaying lower representation at the top of a ranking is categorized as the disadvantaged group. 

\subsection{Model Training and  Parameter Settings} \label{sec:training}
For each dataset listed in Section \ref{sec:datasets},
we consider fairness-unaware models trained without the protected attributes, fairness-unaware models trained reliant on the protected attributes, and fairness-aware models requiring the protected attributes.
Additional information regarding the training of models has been provided in Section~\ref{sec:training} from the supplemental material.

\subsection{Controlled Study: Varying Inference Errors}\label{simulations}
In this first study, we systematically vary the level of  errors in the protected attributes in our ranking scenarios,
mimicking errors commonly encountered in demographic inference. Specifically, we consider three cases: bidirectional errors (\textbf{\smone}), i.e., some disadvantaged candidates are incorrectly inferred as advantaged and vice-versa, and unidirectional errors, where only candidates of one of groups are mistaken by candidates of the other (\textbf{\smtwo} and \textbf{\smthree}). Unidirectional errors capture the situation where candidates whose group could not be inferred are all assigned to either $g_{adv}$ (or $g_{dis}$). We control the error level using a parameter $\epsilon$ defined as the fraction of candidates in each group for which inference is wrong. The results for the unidirectional scenarios are discussed in the supplemental material.

For each of these scenarios, we generate controlled test sets by varying $\epsilon$ from $0\%$ to 
$100\%$ in increments of $10\%$. To reduce the variance in the results, we increase the error levels cumulatively, i.e., all  candidates whose protected attribute was flipped at one error level will retain the wrong value at a higher error level. This process is repeated 5 times with different random seeds for $\epsilon$ from $10\%$ to 
$90\%$. Noting that for $\epsilon$ equal to $0\%$ and $100\%$ the results remain the same, this yields 47 test scenarios for each dataset.
We employ these scenarios as a test benchmark for the approaches previously described in Section \ref{sec:approaches_desc}. 

We compute fairness metrics using ground-truth labels.  We average over all $\epsilon$ values.  

Note that $\epsilon=0\%$ serves as a reference point, as it corresponds to using the ground-truth protected attribute values.

\subsection{Case Studies: Using  Popular  Inference Methods on Real-World Data Sets}\label{cases}

In this second study, we use the  real-world inference services described in Section \ref{sec:services} to obtain demographic attributes for the three real-world datasets introduced in Section \ref{sec:datasets}.

Table \ref{inaccuracies} shows the accuracy of each inference service when provided with candidates' names from our datasets. Each inference service has a subset of candidate items for which it cannot identify a sex.

In such cases, the inference service returns an ``unknown'' value. Gender API has the lowest number of unknown values, 
whereas Namsor has the highest. When considering the accuracy for only the identified items, Namsor performs best. 

\subsubsection*{Handling Unrecognized Names}
Each inference service returned some ``unknowns'', i.e., candidates' sex could not be inferred. Each of these candidates was assigned to the disadvantaged group \dadv{}. This design choice would prevent a disadvantaged candidate to be misclassified as part of \adv{} and thus receive a penalty when fairness-aware strategies are in place. Table~\ref{inaccuracies} shows the number of candidates without inference results. It is important to note that the final error rate changes due to adding the unrecognized names to a group. 

\subsection{Measuring Fairness and Utility}
To calculate the fairness metrics, we use the ground-truth sex attribute to capture true fairness levels. For utility, we use the ground-truth judgment scores.
To analyze the effect of wrong demographic inference on the fairness and utility measures of rankings generated by fairness-aware algorithms, we follow the steps sketched below.

\begin{description}[wide,labelindent=0pt,parsep=0pt,itemsep=5pt]
    \item[Step 1.] We measure the skews (see Definition in Section \ref{skew_calc}) 
    of each group on the ranking directly obtained from judgment scores, i.e., before applying any ranking algorithm. For each dataset, the group that has the lowest skew for all or nearly all positions at the top of the ranking  is defined as the disadvantaged group \dadv{} (see  Table~\ref{tab:dataset}).
    \item[Step 2.] \label{2} 
    We train each ranking model as in Section \ref{sec:training}.
    \item[Step 3.] Lastly, we apply each model to a test set (according to each strategy in Section \ref{sec:approaches_desc}),  and compute fairness \dadexp{} and NDKL and utility NDCG metrics.
  
    In the controlled error rate experiments, we average the metrics over the five variants of the test set generated for each $\epsilon$. 
    
\end{description}

\section{Results and Analysis}\label{sect.results}
We present our results for the 1$^\mathrm{st}$ simulation scenario of the controlled inference experiments in Section \ref{sec-synthetic-study-main}
and  our use cases  
with popular inference services on
 real-world data sets
in Section
\ref{sec-use-cases}. Results for the 2$^\mathrm{nd}$ and 3$^\mathrm{rd}$ simulation scenarios are available in the supplemental material (Figs.~\ref{fig:sims2}, \ref{fig:sims3}).

\subsection{Controlled Inference Error Evaluation}
\label{sec-synthetic-study-main}

\begin{figure*}[ht]
    \centering

    \begin{subfigure}{1\textwidth}
        \includegraphics[width=\linewidth]{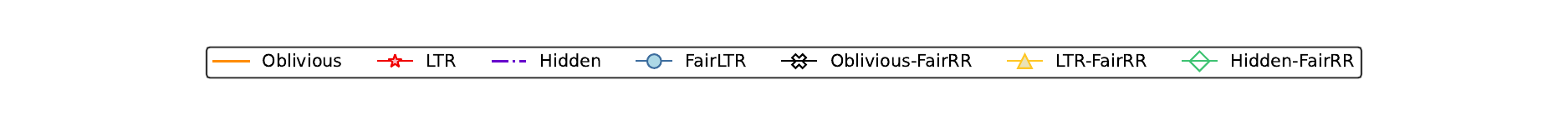}
        
    \end{subfigure}
    \hfill
    \begin{subfigure}{0.24\textwidth}    \includegraphics[width=\linewidth]{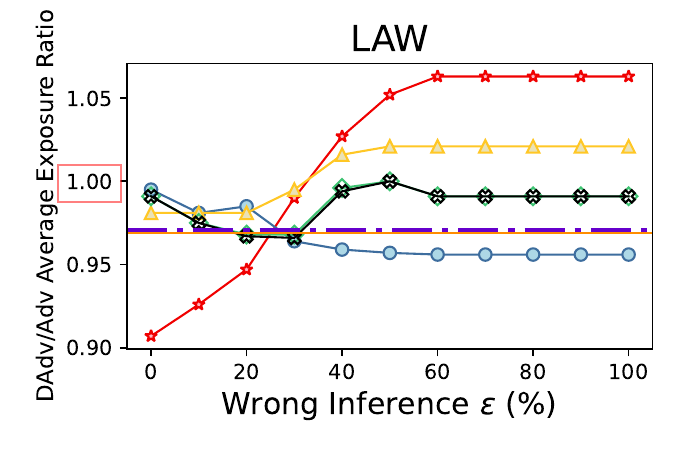}

    \end{subfigure}
    \hfill
    \begin{subfigure}{0.24\textwidth}
        \includegraphics[width=\linewidth]{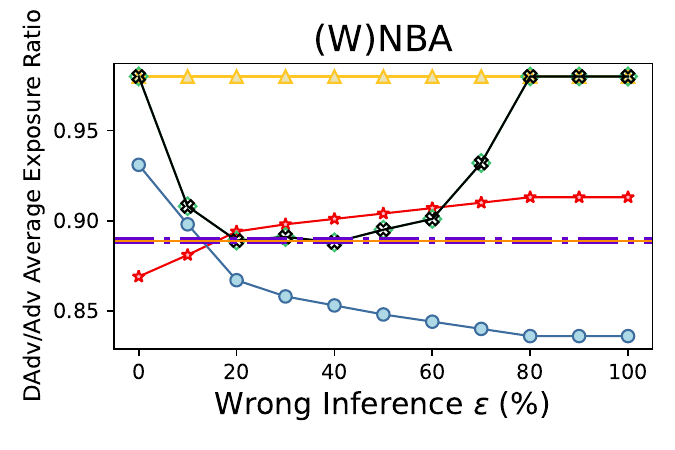}
    \end{subfigure}
    \hfill
    \begin{subfigure}{0.24\textwidth}
        \includegraphics[width=\linewidth]{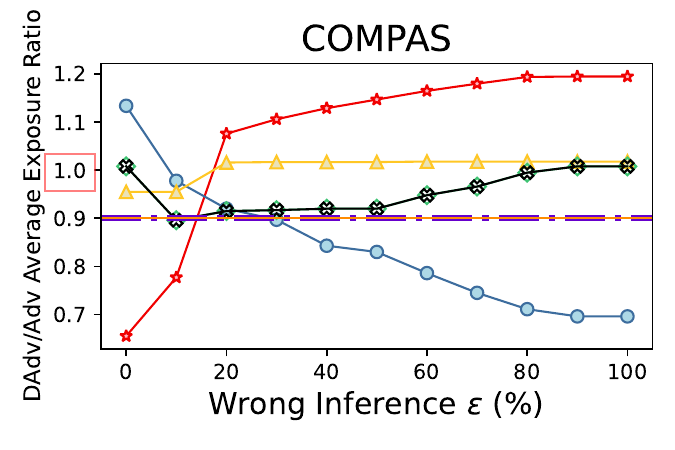}
    \end{subfigure}
    \hfill
    \begin{subfigure}{0.24\textwidth}
        \includegraphics[width=\linewidth]{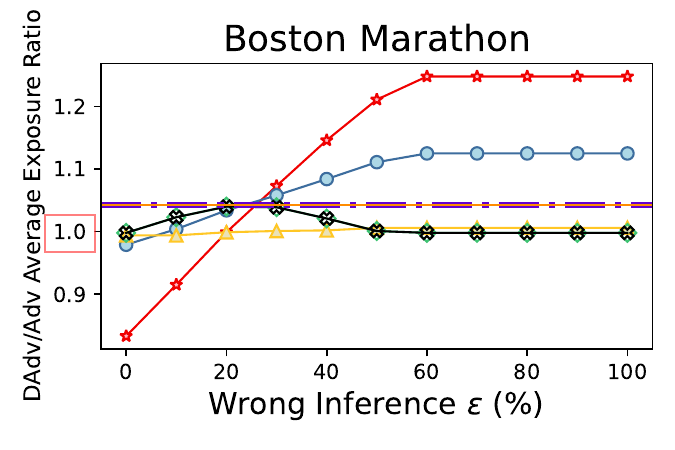}

    \end{subfigure}
    \hfill
    \begin{subfigure}{0.24\textwidth}
        \includegraphics[width=\linewidth]{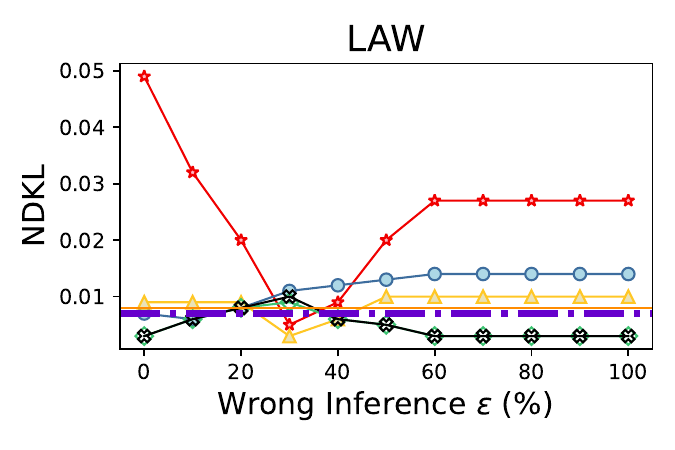}
    \end{subfigure}
    \hfill
    \begin{subfigure}{0.24\textwidth}
        \includegraphics[width=\linewidth]{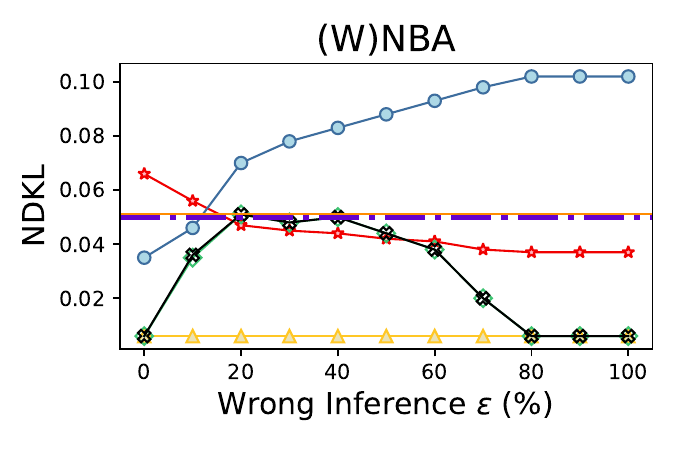}
        
    \end{subfigure}
    \hfill
    \begin{subfigure}{0.24\textwidth}
        \includegraphics[width=\linewidth]{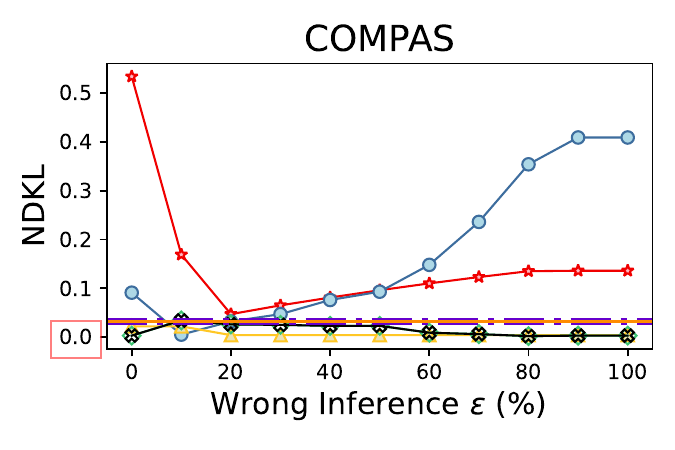}
        
    \end{subfigure}
    \hfill
    \begin{subfigure}{0.24\textwidth}
        {\includegraphics[width=\linewidth]{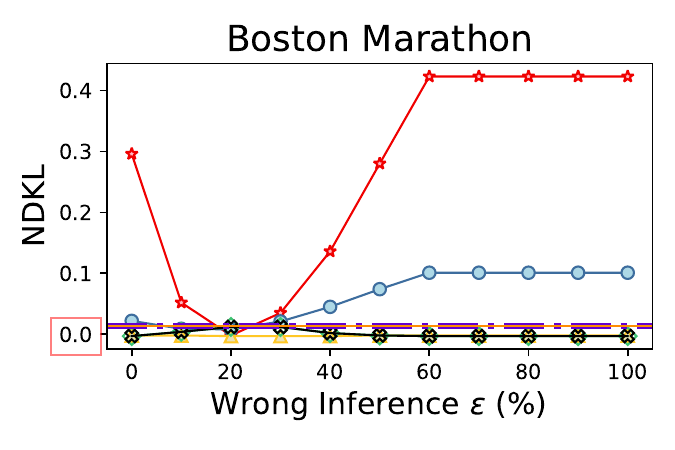}}
    \end{subfigure}   
    \hfill
    
    \begin{subfigure}{0.24\textwidth}
        \includegraphics[width=\linewidth]{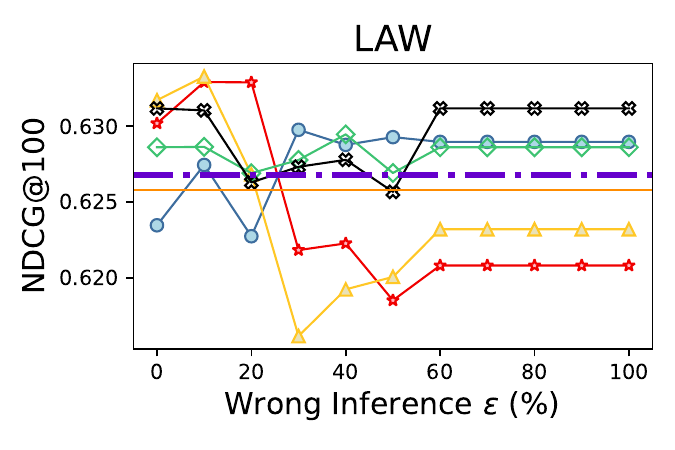}
    \end{subfigure}
    \hfill
    \begin{subfigure}{0.24\textwidth}
        \includegraphics[width=\linewidth]{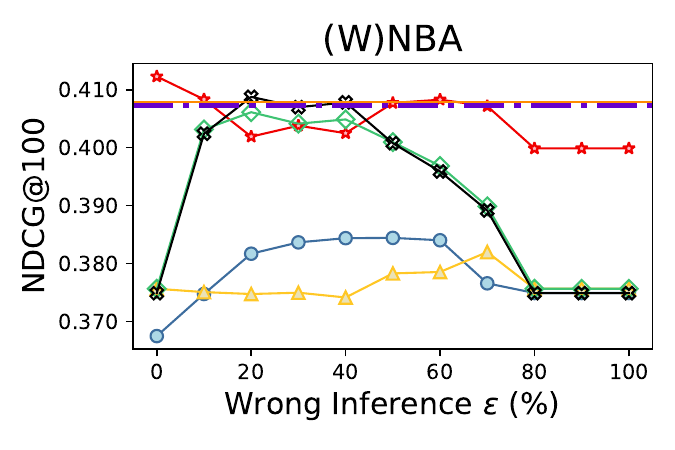}
        
    \end{subfigure}
    \hfill
    \begin{subfigure}{0.24\textwidth}
        \includegraphics[width=\linewidth]{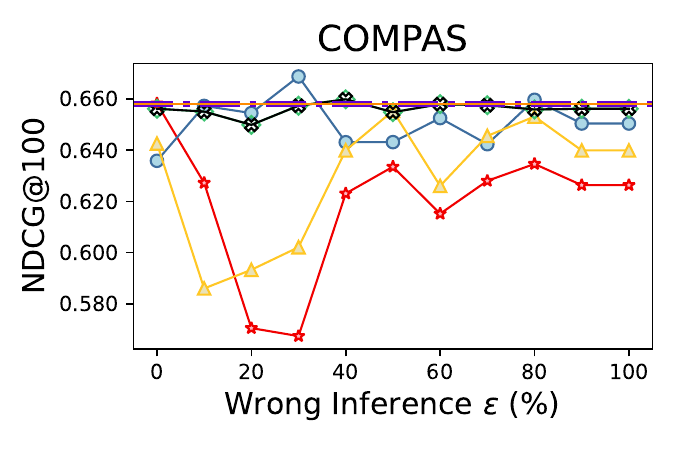}
    \end{subfigure}
    \hfill
    \begin{subfigure}{0.24\textwidth}
        \includegraphics[width=\linewidth]{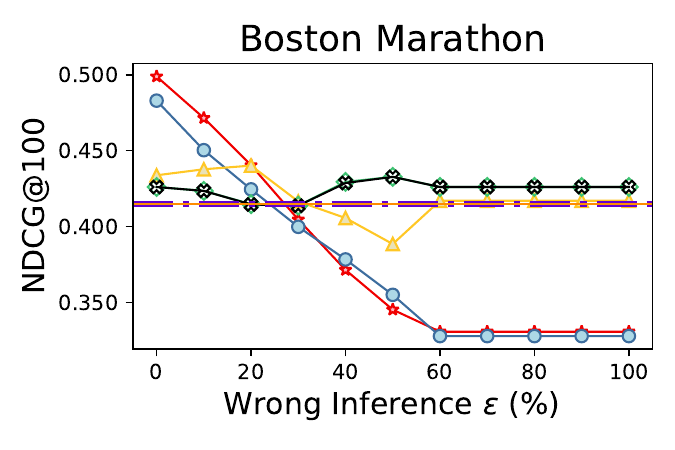}
    \end{subfigure}
    \caption{
    DAdv/Adv Exposure Ratio, NDKL \& NDCG@100 graphs for the 1$^\mathrm{st}$ simulation scenario ($g_{dis}\leftrightarrow g_{adv}$).} Ideal values are highlighted with red boxes on the y-axes whenever visible.
    \label{fig:sims1}
\end{figure*}

Figure~\ref{fig:sims1} shows our results in terms of the fairness and utility measures for the datasets when we have applied controlled inferencing processes to infer protected attributes. For fairness, we present \dadexp{} plots and NDKL plots.
For utility, we present NDCG@100.  

\begin{description}
\item[\algoone{}] \ \
\algoone{} is represented in Figure~\ref{fig:sims1} by a horizontal line, as it is invariant to the inference error $\epsilon$. 
\begin{itemize}[leftmargin=1pt, align=left, labelindent=0pt]  
        \item \algoone{} consistently has \dadexp{} farther from $1.0$ than the fair interventions for up to 15-20\% error across all the datasets. This is expected, as the model does not try explicitly optimize fairness. 
        This shows that if the error is small enough, fairness interventions are a sufficient and safe option for practitioners (as opposed to using an oblivious model). Moreover, it consistently yields similar \dadexp{} values to \algothree{}, showing that hiding the protected attributes during testing would yield fairness levels to those when the protected attribute is not used during training. 
        \item  
        Opting for \algoone{} when subject to possible errors in the protected attribute values also proves to be a better option than choosing \algotwo{} in terms of \dadexp{} for up to 15 - 20\% error rate. This is expected, as including the protected attributes in an already biased ranking only reinforces the bias.
        This implies that, disregarding the protected attribute during training can improve fairness even though it is not explicitly optimize.
        \item For NDKL, \algoone{}'s value is similar to those of the reranking fairness interventions' across different error rates except for in the (W)NBA dataset, where \algoone{} does better.
        \item For NDCG, in cases where the \dadexp{} were lower than those of the fairness strategies, \algoone{} had higher NDCG values and vice versa (except for the LAW dataset). This is as expected due to the fairness-utility trade-off.
    \end{itemize}
    
\item [\algotwo{}] \
\begin{itemize}[leftmargin=1pt, align=left, labelindent=0pt]
    \item For \algotwo{}, the \dadexp{} rises with the inference error across all datasets until it eventually converges. This  is in line with our expectations, as errors in demographic attributes lead the model to misidentify the disadvantaged individuals as advantaged, resulting in a heightened degree of preference towards the disadvantaged group, \dadv.
    \item Interestingly, while fairness is not an inherent objective of the model, group fairness emerges as a byproduct of erroneous inferences throughout the rankings. However, it is worth noting that for $\epsilon \geq 30\%$, the initial disparity is reversed and the advantaged group gets less exposure. The only exception is (W)NBA, where the effect of the sex coefficient is relatively small (as seen by the small difference in \dadexp{} between \algotwo{} at $\epsilon=0\%$ and \algothree{}).
    
    \item NDKL decreases (increased fairness) with inference error, however, there is a turning point after which the original advantaged group becomes severely disadvantaged. This is in line with results for \dadexp.

    \item  NDCG tends to decrease with inference noise, as advantaged (resp.\ disadvantaged) candidates move down (resp.\ up) the ranking. This effect is more pronounced in the COMPAS and Boston Marathon  data sets because the coefficient associated with the protected attribute has a large magnitude. 
\end{itemize}

\item [\algothree{}] Represented in Figure~\ref{fig:sims1} by a horizontal dashed line, it is invariant to the inference error $\epsilon$. 
\begin{itemize}[leftmargin=1pt, align=left, labelindent=0pt]
    
    \item  In the cases where the inference error rate is up to 15-20\%, this method is outperformed by all the fairness strategies, which yield \dadexp{} closer to 1.0. In contrast, for higher error rates, the effectiveness of the fairness strategies drops significantly. 
    \item We observe that the NDKL is consistently outperformed by all the fairness strategies for up to 15-20\% inference error rates, which is similar to the \dadexp{} staying close to 1.0 in our results. 
    \item The NDCG of \algothree{} tends to be higher (i.e., better) than those of the fair strategies, because it does not attempt to compromise utility and fairness.
\end{itemize}

\item [\algofour{}] \ 
\begin{itemize} [leftmargin=1pt, align=left, labelindent=0pt]
    \item For the fair LTR fairness strategy, \algofour{}, \dadexp{} decreases (decreased fairness) with the error rate across the first three datasets  in all scenarios. When candidates in \dadv{} are incorrectly identified as part of \adv, they do not get the score boost the model would have given them otherwise, therefore decreasing their exposure. The only exception, Boston Marathon, is easily explained by inspecting the model coefficients: although smaller in magnitude than in \algotwo{}, \algofour{} still yields a negative coefficient associated with being part of the disadvantaged group. Hence, as the inference errors increase, disadvantaged candidates  benefit. 
    \item 
    NDKL tends to always increase (decreased fairness) and then converge. This is because NDKL summarizes unfairness across groups -- i.e., its value does not reveal which group is disadvantaged. Whereas \dadexp~ quantifies which groups is over- or under-advantaged. 

\item NDCG tends to increase as advantaged (resp. disadvantaged) candidates get a bonus (resp. penalty) in score due to being incorrectly inferred as part of the other group. 
\end{itemize}

\item [\algofive{}] \ 
\begin{itemize}[leftmargin=1pt, align=left, labelindent=0pt]

\item For \algofive{}, there is not much variation in the \dadexp{} on the Law and Boston Marathon datasets. The value initially  reduces in the fairness value (away from 1.0), but eventually returns to its inital value. This is due to the fact that, from the DetConstSort's perspective, the group proportions between $g_{dis}$ and $g_{adv}$ have been swapped. For instance, in the (W)NBA case, at $\epsilon=100\%$, DetConstSort perceives 78\% of the population as female (i.e.,  the inverse of Table~\ref{tab:dataset}).
This will ensure that this proportion is present at any top $k$ positions in the ranking. However, since the sex attribute values are completely swapped, but candidates' scores remain the same,  DetConstSort will put the same candidates in the same positions as had the inferred attributes been $100\%$ accurate. Thus it recovers the exact same ranking as in $\epsilon=0$. 

\item Consequently, NDKL does not change much with error for \algofive{}. Even in (W)NBA, the NDKL increase is relatively small: its maximum is 0.05. 

\item NDCG does not vary much for LAW, COMPAS and Boston Marathon. 
The NDKL returns to the same value as for $\epsilon=0\%$ as the error approaches $100\%$, since the two rankings are identical.

\end{itemize}

\item [\algosix{}] \ 
\begin{itemize} [leftmargin=1pt, align=left, labelindent=0pt]
    \item Surprisingly, for \algosix{}, \dadexp{} does not exhibit much variation with inference error in any dataset. After careful investigation, we found that this is because the increase/decrease in score caused by providing the wrong protected attribute to Listnet is often counteracted by DetConstSort when the perceived group proportions change, as we explain next. For instance, a candidate $c$ from $g_{dis}$ perceived by Listnet as part of $g_{adv}$ gets a bonus in their score. 
Yet, these inference errors also lead to a small increase in the proportion of the advantaged group.

\item The NDCG decreases and then converges across the  increases error level for the LAW and Boston Marathon datasets, it is for stable for the (W)NBA dataset and less stable for COMPAS.
\end{itemize}

\item [\algoseven{}] \

The results are identical to \algofive{} for the \dadexp{} and NDKL. This is expected, following the similarity between \algoone
{} and \algothree{}. Slight differences are however observed in the NDCG values for the LAW and (W)NBA datasets.
\end{description}

\subsection{Real World Data Set Use Cases with SOTA Inference
Techniques}
\label{sec-use-cases}

\begin{figure*}[ht]
\centering
\begin{subfigure}{.95\textwidth}
\includegraphics[width=1\linewidth]{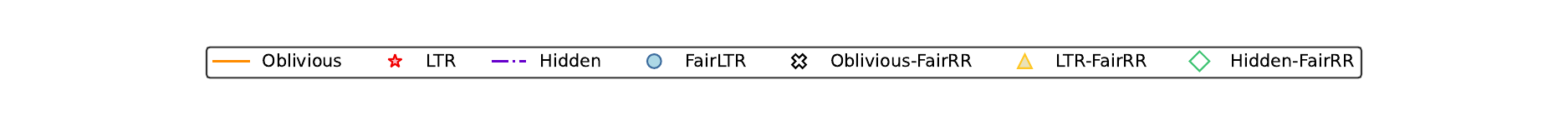}
\end{subfigure}
\begin{subfigure}{0.26\linewidth}
\includegraphics[width=1\linewidth]{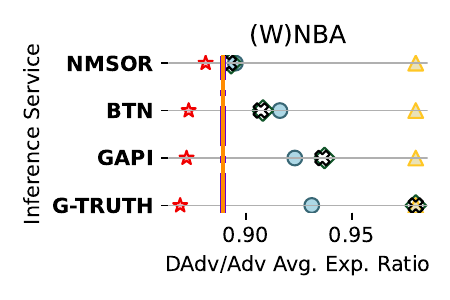}
\end{subfigure}
\begin{subfigure}{0.26\linewidth}
\includegraphics[width=1\linewidth]{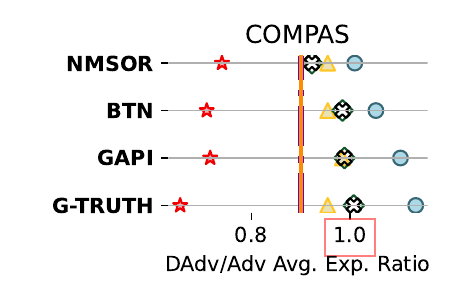}
\end{subfigure}
\begin{subfigure}{0.26\linewidth}
\includegraphics[width=1\linewidth]{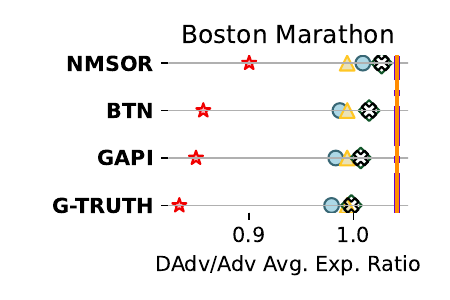}
\end{subfigure}
\begin{subfigure}{0.26\linewidth}
\includegraphics[width=1\linewidth]{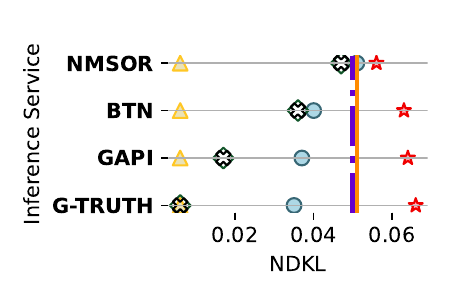}
\end{subfigure}
\begin{subfigure}{0.26\linewidth}
\includegraphics[width=1\linewidth]{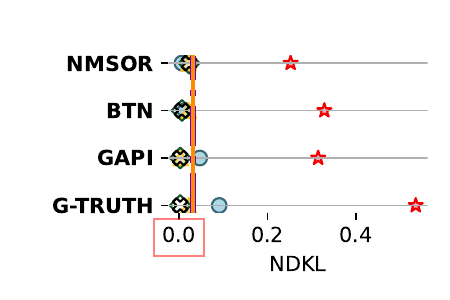}
\end{subfigure}
\begin{subfigure}{0.26\linewidth}
\includegraphics[width=1\linewidth]{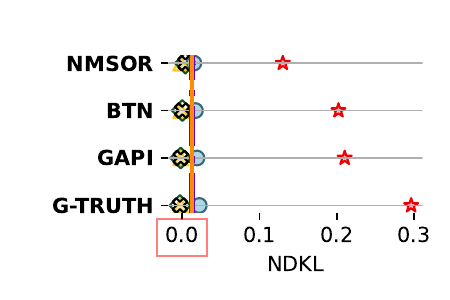}
\end{subfigure}
\begin{subfigure}{0.26\linewidth}
\includegraphics[width=1\linewidth]{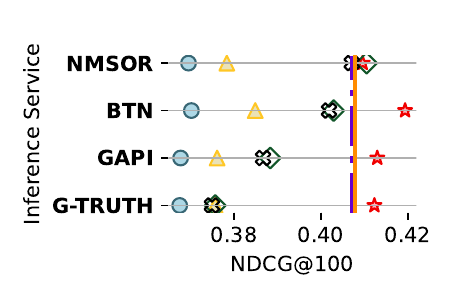}
\end{subfigure}
\begin{subfigure}{0.26\linewidth}
\includegraphics[width=1\linewidth]{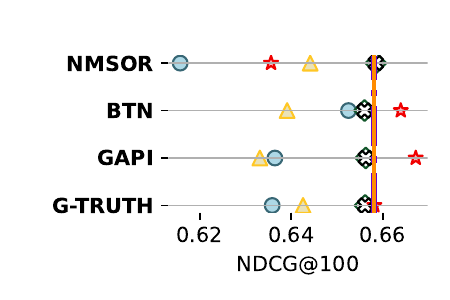}
\end{subfigure}
\begin{subfigure}{0.26\linewidth}
\includegraphics[width=1\linewidth]{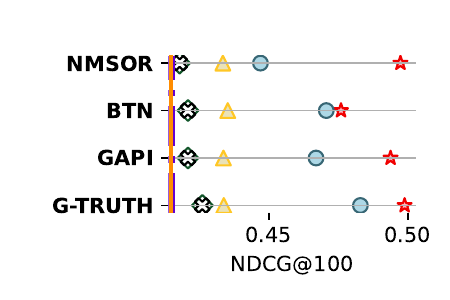}
\end{subfigure}

    \caption{DAdv/Adv Exposure Ratio, NDKL and NDCG@100 graphs for all inference services and LTR strategies.}
    \label{fig:case}

\end{figure*}

As explained in Section \ref{sec:services}, we assign the candidates whose protected attribute value were unknown to the disadvantaged group.
For the ease of exposition, we sort services \emph{in increasing order of error rate}: Gender API (GAPI), Behind The Name (BTN) and Namsor (NMSOR). We add a  case for  ground-truth protected demographics (0\% error rate), 
referred to as G-TRUTH in Figure 
\ref{fig:case}.
In general, we observe similar behavior as in the error simulation studies.

\begin{description}

\item[\algoone{}] \ It yields \dadexp{} values farther away from 1.0 than all the fair interventions in the G-TRUTH results. Yet, it is closer to 1.0 than \algotwo{}. 

\item[\algotwo{}] We observe that the \dadexp{} values slowly increases from the least to the most inaccurate service, exhibiting a slight variation with smaller errors (6\% for GAPI for (W)NBA) and a larger variation with larger errors (47\% for NMSOR for (W)NBA). This is consistent with our results in Figure~\ref{fig:sims1}). NDKL and NDCG results are also in line with our controlled experiments. 

\item[\algothree{}] In terms of \dadexp{}, fairness strategies yield values close to 1.0 on Boston Marathon, whereas \algothree{} ends up overexposing the disadvantaged group. This is considered unfair as it deviates from the 
ideal \dadexp{} value. Therefore, \algothree{} is outperformed by the fairness strategies in terms of fairness. Conversely, regarding NDCG, \algothree{} matches or exceeds the performance of fairness strategies on (W)NBA and COMPAS. Due to the special nature of the Boston Marathon dataset (where the disadvantaged group is overexposed by \algothree{}), 
\algothree{} leads to the lowest utility among all strategies.

\item[\algofour{}] Fairness decreases (\dadexp{} deviates further from 1.0) observe with an increase in error rates in the (W)NBA dataset and in Boston Marathon. COMPAS is an exceptional case since \algofour{} overexposes the disadvantaged group, so an increase in error brings \dadexp{} closer to 1.0. NDKL mirrors the same fairness trends as in the \dadexp{} values. For COMPAS and Boston Marathon, NDCG are the smallest values for the most inaccurate service (NMSOR). 

\item[\algofive{}] The larger the error in inference, the father away from 1.0 the \dadexp{} value is. For NDKL, the deviation from the perfect representation as the error increases is not clearly seen except in  (W)NBA dataset. NDCG values reflect the fairness-utility trade-off; higher NDCG values for lower \dadexp{} values and vice-versa. 

\item[\algosix{}] \dadexp{} again stays relatively constant across the inference services. In terms of NDCG, this strategy achieves consistent performance across different services.

\item[\algoseven{}] \dadexp{}  stays relatively constant with increases in inference error rate, corroborating the results from the controlled experiments. Observed trends for NDCG are also as expected: they are somewhat invariant to inference errors, exhibiting only a minor increase with error in (W)NBA dataset.

\end{description}
\section{Discussion: Insights and Take-Aways}
Our paper focuses on how errors in inferring the protected attribute may
affect the fairness and utility metrics of rankings produced by a wide spectrum of
alternate LTR-based ranking systems.
These methods can instill fairness at one of two possible stages: as part of a Fair LTR model  or a posteriori by using a Fair ReRanker. We investigated the feasibility and implications of inferring or neglecting protected attributes when they are not available.
We conclude:

\begin{itemize}
    \item In scenarios characterized by high inference error rates, fairness models that require protected attributes, while still effective in increasing the disadvantaged group exposure, may inadvertently lead
    to the advantaged group having lower exposure than the former
    
    (e.g, Boston Marathon in Fig.~\ref{fig:sims1}).
    
    \item The impact of inference errors on fairness varies 
    depending on the model employed. Notably, fair LTR models such as DELTR exhibit distinct behavior compared to fairness-unaware models like ListNet followed by a Fair ReRanker.  The latter demonstrates superior capability in maintaining fairness when faced with higher levels of inference errors.
    While LTR boosts the scores of candidates incorrectly inferred as advantaged, DetConstSort prevents overexposure of this group by enforcing that both candidates inferred as advantaged or disadvantaged are proportionally represented at all ranking cutoffs.
    \item  We studied 
    whether to hide or infer protected features
    if they are not readily available. 
For the \algotwo{} strategy, we  observated that wrong inference can improve fairness, as candidate items from the disadvantaged group typically get a better rank when misclassified.
Nonetheless, the takeaway is that inference methods are often unpredictable, and their use should be heavily monitored, if not discouraged.
 Both our experiments with controlled inference error rates and real world use cases with state-of-the-art
inference techniques corroborate that finding.
\item \algofour{} strategies like DELTR  maintain fairness even with imperfect inference, with better fairness achieved as inference accuracy increases. Conversely, strategies that do not depend on inferring protected attributes, like \algothree{} may provide lower fairness metrics overall. 
 
\end{itemize}
Our research emphasizes the dangers of demographic inference for practitioners along with other important insights.

\paragraph{Study Limitations and Future Work.}

While our
study focuses on gender,
other demographics like race or religion along with their inferencing services also are needed
  to reach a general recommendation for practitioners.   Demographic attributes can be multi-valued, while our investigation,
similar to  prior work \cite{ghosh2021fair},
focused the core binary scenario.

\section*{Conclusion}
The absence of
demographic features may affect the effective functioning of fair ranking systems.
Thus, practitioners may   attempt to overcome this by either promoting fairness despite their
absence  or turning to demographic inference
tools to attempt to infer them. Our study sheds
some light on this decision,
namely, we find that
 if this inference is deemed necessary and inevitable, fair re-ranking solutions serve as a more resilient alternative compared to Fair LTR  solutions.

\section*{Acknowledgments}

This work is supported in part by NSF grants IIS 2007932,
CNS-1852498, IIS-1910880, CSSI-2103832,   and AAUW.
Thank you to undergraduate REU/MQP researchers Pietrick, Romportl, Smith, Tessier, Vadlamudi, and Venkataraman for contributions to early stages of this project.
Thanks Daisy research group for valuable feedback.

\section*{Ethics Statement}

\paragraph{Ethical Considerations}

We worked with datasets, AI models, and inference tools that are all publicly available. 

As such, we adhered to all licensing agreements and download requirements related to their use.
Our experimental approach is explained in Section \ref{sec:methodology}.

For the datasets, we do not display any information about these datasets at the granularity of individuals in our work -
only at the aggregated level of demographic groups. Licensing details about the models and inferencing tools are found in Table \ref{tab:license}. 
\begin{table*}[ht]
    \centering  
    \begin{tabular}{l|l}
       \textbf{Model/Tool}  & \textbf{License Details} \\ \midrule
       DELTR  & \url{https://github.com/fair-search/fairsearch-deltr-python/blob/master/LICENSE}/pp\\
       
       GenderAPI & \url{https://gender-api.com/en/terms-of-use} \\
       Behind The Name & \url{https://www.behindthename.com/info/terms}\\
       Namsor & \url{https://namsor.app/terms-and-conditions/}

    \end{tabular}
    \caption{License details}
    
\label{tab:license}
\end{table*}
All source code we produced and experimental artifacts of our experimental study will be made available on \textit{github},
upon acceptance of this paper.

\paragraph{Adverse Impacts}
While our research finds that using inaccurate demographic inference tools is typically inadvisable, we are mindful that fairness is contextual to the problem, tools, and task at hand. Practitioners should continuously monitor the fairness performance of ranking models and as well as assess whether using inference methods in place of demographic data has unintended consequences. 

Understanding the ramifications of using demographic inference tools, as done in our work, should in no way be construed as endorsing such practices. Rather, with this work, we aim to uncover the potential drawbacks and benefits that these practices may cause so that practitioners and policy makers can make informed decisions.

\paragraph{Researcher Positionality}
As computer scientists and data scientists by training,   we are influenced by our perspectives on how we approach research problems and how we develop and/or study computational solutions to fairness problems 
in our society related to the introduction of digital tools.

\bibliography{aaai24}

\begin{thebibliography}{39}
\providecommand{\natexlab}[1]{#1}

\bibitem[{BehindTheName(1996)}]{btn}
BehindTheName. 1996.
\newblock \url{https://www.behindthename.com/}.
\newblock Last Accessed: January 21, 2024.

\bibitem[{Bogen, Rieke, and Ahmed(2020)}]{bogen2020awareness}
Bogen, M.; Rieke, A.; and Ahmed, S. 2020.
\newblock Awareness in practice: tensions in access to sensitive attribute data for antidiscrimination.
\newblock In \emph{ACM FAT*}, 492--500.

\bibitem[{Cao et~al.(2007)Cao, Qin, Liu, Tsai, and Li}]{cao2007learning}
Cao, Z.; Qin, T.; Liu, T.-Y.; Tsai, M.-F.; and Li, H. 2007.
\newblock Learning to rank: from pairwise approach to listwise approach.
\newblock In \emph{ICML}, 129--136.

\bibitem[{Celis et~al.(2021)Celis, Huang, Keswani, and Vishnoi}]{celis2021fair}
Celis, L.~E.; Huang, L.; Keswani, V.; and Vishnoi, N.~K. 2021.
\newblock Fair classification with noisy protected attributes: A framework with provable guarantees.
\newblock In \emph{ICML}, 1349--1361.

\bibitem[{Cheng et~al.(2009)Cheng, Chen, Chandramouli, and Subbalakshmi}]{emailgender}
Cheng, N.; Chen, X.; Chandramouli, R.; and Subbalakshmi, K. 2009.
\newblock Gender identification from e-mails.
\newblock In \emph{IEEE SSCI}, 154--158.

\bibitem[{{Council of the European Union}(2016)}]{EUdataregulations2016}
{Council of the European Union}. 2016.
\newblock Regulation ({EU}) 2016/679 of the {European} {Parliament} and of the {Council}.

\bibitem[{Dwork et~al.(2012)Dwork, Hardt, Pitassi, Reingold, and Zemel}]{dwork2012fairness}
Dwork, C.; Hardt, M.; Pitassi, T.; Reingold, O.; and Zemel, R. 2012.
\newblock Fairness through awareness.
\newblock In \emph{ITCS}, 214--226.

\bibitem[{{eCFR}(2024)}]{ecfr}
{eCFR}. 2024.
\newblock {Electronic Code of Federal Regulations}.
\newblock \url{https://www.ecfr.gov/current/title-12/chapter-X/part-1002/subpart-A/section-1002.5#p-1002.5(b)}.
\newblock May 9, 2024.

\bibitem[{Ekstrand et~al.(2021)Ekstrand, Das, Burke, and Diaz}]{ekstrand2021fairness}
Ekstrand, M.~D.; Das, A.; Burke, R.; and Diaz, F. 2021.
\newblock Fairness and discrimination in information access systems.
\newblock \emph{arXiv preprint arXiv:2105.05779}.

\bibitem[{Fink, Kopecky, and Morawski(2012)}]{socialmediainferring}
Fink, C.; Kopecky, J.; and Morawski, M. 2012.
\newblock Inferring gender from the content of tweets: A region specific example.
\newblock In \emph{AAAI ICWSM}, 459--462.

\bibitem[{{Gender-API}(n.d.)}]{genderapi}
{Gender-API}. n.d.
\newblock {Gender-API (Version 2)}.
\newblock \url{https://gender-api.com}.
\newblock Last Accessed: January 21, 2024.

\bibitem[{Geyik, Ambler, and Kenthapadi(2019)}]{geyik}
Geyik, S.~C.; Ambler, S.; and Kenthapadi, K. 2019.
\newblock Fairness-aware ranking in search \& recommendation systems with application to \uppercase{L}inkedIn talent search.
\newblock In \emph{ACM SIGKDD}, 2221--2231.

\bibitem[{Ghazimatin et~al.(2022)Ghazimatin, Kleindessner, Russell, Abedjan, and Golebiowski}]{ghazimatin2022measuring}
Ghazimatin, A.; Kleindessner, M.; Russell, C.; Abedjan, Z.; and Golebiowski, J. 2022.
\newblock Measuring fairness of rankings under noisy sensitive information.
\newblock In \emph{ACM FAccT}, 2263--2279.

\bibitem[{Ghosh, Dutt, and Wilson(2021)}]{ghosh2021fair}
Ghosh, A.; Dutt, R.; and Wilson, C. 2021.
\newblock When fair ranking meets uncertain inference.
\newblock In \emph{ACM SIGIR 2021}, 1033--1043.

\bibitem[{Ghosh, Kvitca, and Wilson(2023)}]{ghosh2023fair}
Ghosh, A.; Kvitca, P.; and Wilson, C. 2023.
\newblock When Fair Classification Meets Noisy Protected Attributes.
\newblock In \emph{AAAI/ACM AIES}, 679--690.

\bibitem[{Goel et~al.(2021)Goel, Amayuelas, Deshpande, and Sharma}]{goel2021importance}
Goel, N.; Amayuelas, A.; Deshpande, A.; and Sharma, A. 2021.
\newblock The importance of modeling data missingness in algorithmic fairness: A causal perspective.
\newblock In \emph{Proceedings of the AAAI Conference on Artificial Intelligence}, 7564--7573.

\bibitem[{Hashimoto et~al.(2018)Hashimoto, Srivastava, Namkoong, and Liang}]{pmlr-v80-hashimoto18a}
Hashimoto, T.; Srivastava, M.; Namkoong, H.; and Liang, P. 2018.
\newblock Fairness Without Demographics in Repeated Loss Minimization.
\newblock In \emph{ICML}, 1929--1938.

\bibitem[{Holstein et~al.(2018)Holstein, Vaughan, Daum{\'e}, Dud{\'i}k, and Wallach}]{Holstein2018ImprovingFI}
Holstein, K.; Vaughan, J.~W.; Daum{\'e}, H.; Dud{\'i}k, M.; and Wallach, H.~M. 2018.
\newblock Improving Fairness in Machine Learning Systems: What Do Industry Practitioners Need?
\newblock In \emph{ACM CHI}.

\bibitem[{J{\"a}rvelin and Kek{\"a}l{\"a}inen(2002)}]{jarvelin2002cumulated}
J{\"a}rvelin, K.; and Kek{\"a}l{\"a}inen, J. 2002.
\newblock Cumulated gain-based evaluation of IR techniques.
\newblock \emph{ACM TOIS}, 20(4): 422--446.

\bibitem[{Jillson(2021)}]{jillson2021aiming}
Jillson, E. 2021.
\newblock Aiming for truth, fairness, and equity in your company’s use of AI.
\newblock \emph{Federal Trade Commission}.

\bibitem[{K{\i}rnap et~al.(2021)K{\i}rnap, Diaz, Biega, Ekstrand, Carterette, and Yilmaz}]{kirnap2021estimation}
K{\i}rnap, {\"O}.; Diaz, F.; Biega, A.; Ekstrand, M.; Carterette, B.; and Yilmaz, E. 2021.
\newblock Estimation of fair ranking metrics with incomplete judgments.
\newblock In \emph{The Web Conference}, 1065--1075.

\bibitem[{Lahoti et~al.(2020)Lahoti, Beutel, Chen, Lee, Prost, Thain, Wang, and Chi}]{Lahoti2020FairnessWD}
Lahoti, P.; Beutel, A.; Chen, J.; Lee, K.; Prost, F.; Thain, N.; Wang, X.; and Chi, E.~H. 2020.
\newblock Fairness without Demographics through Adversarially Reweighted Learning.
\newblock \emph{ArXiv}, abs/2006.13114.

\bibitem[{Larson et~al.(2016)Larson, Mattu, Kirchner, and Angwin}]{ProPublica}
Larson, J.; Mattu, S.; Kirchner, L.; and Angwin, J. 2016.
\newblock How we analyzed the \uppercase{COMPAS} Recidivism Algorithm.

\bibitem[{Li et~al.(2021)Li, Chen, Fu, Ge, and Zhang}]{li2021user}
Li, Y.; Chen, H.; Fu, Z.; Ge, Y.; and Zhang, Y. 2021.
\newblock User-oriented fairness in recommendation.
\newblock In \emph{Proceedings of the Web Conference 2021}, 624--632.

\bibitem[{Li et~al.(2022)Li, Chen, Xu, Ge, Tan, Liu, and Zhang}]{li2022fairness}
Li, Y.; Chen, H.; Xu, S.; Ge, Y.; Tan, J.; Liu, S.; and Zhang, Y. 2022.
\newblock Fairness in recommendation: A survey.
\newblock \emph{arXiv preprint arXiv:2205.13619}.

\bibitem[{Mehrotra and Vishnoi(2022)}]{mehrotra2022fair}
Mehrotra, A.; and Vishnoi, N. 2022.
\newblock Fair ranking with noisy protected attributes.
\newblock In \emph{NeurIPS}, 31711--31725.

\bibitem[{Mozannar, Ohannessian, and Srebro(2020)}]{mozannar2020fair}
Mozannar, H.; Ohannessian, M.; and Srebro, N. 2020.
\newblock Fair learning with private demographic data.
\newblock In \emph{ICML}, 7066--7075.

\bibitem[{Namesor(n.d.)}]{nmsor}
Namesor. n.d.
\newblock https://v2.namsor.com/NamSorAPIv2/api2/json/genderBatch.
\newblock Last Accessed: January 21, 2024.

\bibitem[{Noriega-Campero et~al.(2019)Noriega-Campero, Bakker, Garcia-Bulle, and Pentland}]{noriega2019active}
Noriega-Campero, A.; Bakker, M.~A.; Garcia-Bulle, B.; and Pentland, A. 2019.
\newblock Active fairness in algorithmic decision making.
\newblock In \emph{Proceedings of the 2019 AAAI/ACM Conference on AI, Ethics, and Society}, 77--83.

\bibitem[{Patro et~al.(2022)Patro, Porcaro, Mitchell, Zhang, Zehlike, and Garg}]{fairpatro22}
Patro, G.~K.; Porcaro, L.; Mitchell, L.; Zhang, Q.; Zehlike, M.; and Garg, N. 2022.
\newblock Fair Ranking: A Critical Review, Challenges, and Future Directions.
\newblock In \emph{ACM FAccT}, 1929–1942.
\newblock ISBN 9781450393522.

\bibitem[{Santamar{\'\i}a and Mihaljevi{\'c}(2018)}]{accuracycomparison}
Santamar{\'\i}a, L.; and Mihaljevi{\'c}, H. 2018.
\newblock Comparison and benchmark of name-to-gender inference services.
\newblock \emph{PeerJ Computer Science}, 4: e156.

\bibitem[{Singh and Joachims(2018)}]{singh2018fairness}
Singh, A.; and Joachims, T. 2018.
\newblock Fairness of exposure in rankings.
\newblock In \emph{ACM SIGKDD}, 2219--2228.

\bibitem[{Singh and Joachims(2019)}]{singh2019policy}
Singh, A.; and Joachims, T. 2019.
\newblock Policy learning for fairness in ranking.
\newblock \emph{Advances in neural information processing systems}, 32.

\bibitem[{Wang et~al.(2020)Wang, Guo, Narasimhan, Cotter, Gupta, and Jordan}]{wang2020robust}
Wang, S.; Guo, W.; Narasimhan, H.; Cotter, A.; Gupta, M.; and Jordan, M. 2020.
\newblock Robust optimization for fairness with noisy protected groups.
\newblock In \emph{NeurIPS}, 5190--5203.

\bibitem[{Wang, Tao, and Fang(2022)}]{wang2022meta}
Wang, Y.; Tao, Z.; and Fang, Y. 2022.
\newblock A Meta-learning Approach to Fair Ranking.
\newblock In \emph{ACM SIGIR}, 2539--2544.

\bibitem[{Wightman(1998)}]{wightman1998lsac}
Wightman, L.~F. 1998.
\newblock LSAC National Longitudinal Bar Passage Study. LSAC Research Report Series.
\newblock \emph{ERIC}.

\bibitem[{Zehlike et~al.(2017)Zehlike, Bonchi, Castillo, Hajian, Megahed, and Baeza-Yates}]{zehlike2017fa}
Zehlike, M.; Bonchi, F.; Castillo, C.; Hajian, S.; Megahed, M.; and Baeza-Yates, R. 2017.
\newblock Fa* ir: A fair top-k ranking algorithm.
\newblock In \emph{Proceedings of the 2017 ACM on Conference on Information and Knowledge Management}, 1569--1578.

\bibitem[{Zehlike and Castillo(2020)}]{DELTR}
Zehlike, M.; and Castillo, C. 2020.
\newblock Reducing disparate exposure in ranking: A learning to rank approach.
\newblock In \emph{The Web Conference}, 2849--2855.

\bibitem[{Zhang and Long(2021)}]{zhang2021assessing}
Zhang, Y.; and Long, Q. 2021.
\newblock Assessing fairness in the presence of missing data.
\newblock \emph{Advances in neural information processing systems}, 34: 16007--16019.

\end{thebibliography}

\appendix

\section{Additional Results}

\begin{figure*}[ht]
\begin{subfigure}{0.24\linewidth}
\includegraphics[width=1\textwidth]{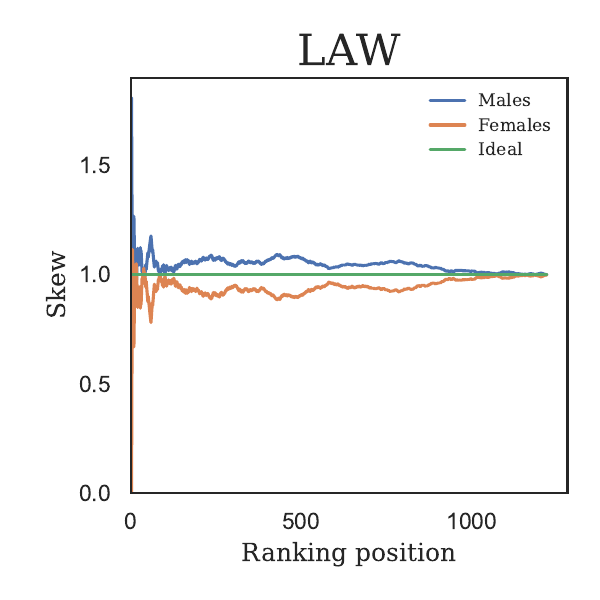}
\end{subfigure}
\begin{subfigure}{0.24\linewidth}
\includegraphics[width=1\textwidth]{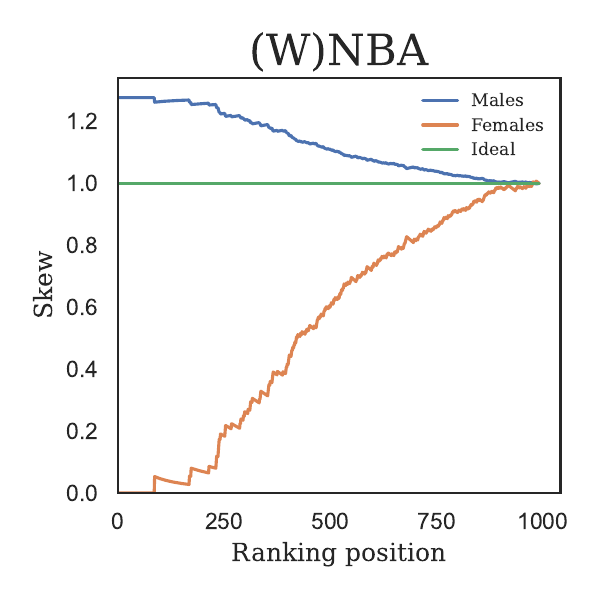}
\end{subfigure}
\begin{subfigure}{0.24\linewidth}
\includegraphics[width=1\textwidth]{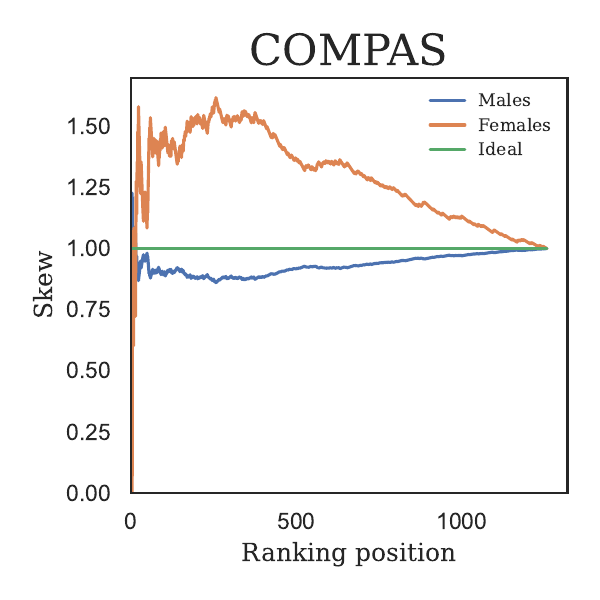}
\end{subfigure}
\begin{subfigure}{0.24\linewidth}
\includegraphics[width=1\textwidth]{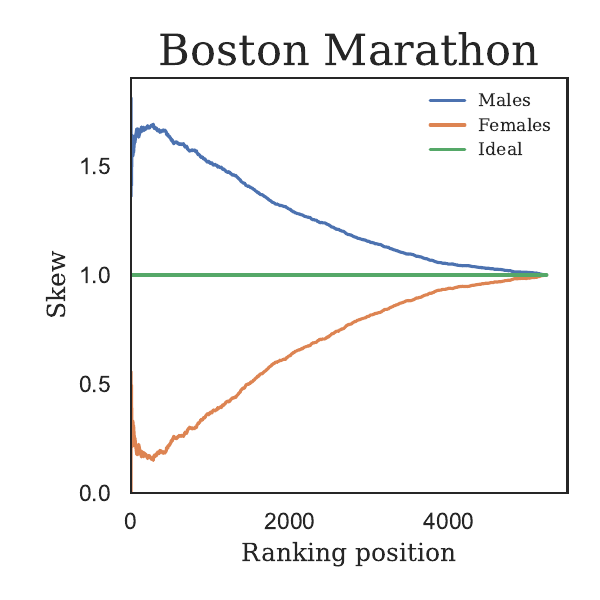}
\end{subfigure}

\caption{Skew graphs of test datasets. The group with skews below the green line is the disadvantaged group.}
\label{fig:skew_graphs}

\end{figure*}

\subsection{NDCG Results}
\label{extra_ndcg}
\begin{figure*}[ht]
    \centering

    \begin{subfigure}{1\textwidth}
        \includegraphics[width=\linewidth]{figures/Simulations/legend_synth.pdf}
        
    \end{subfigure}
    \hfill
    \begin{subfigure}{0.24\textwidth}
        \includegraphics[width=\linewidth]{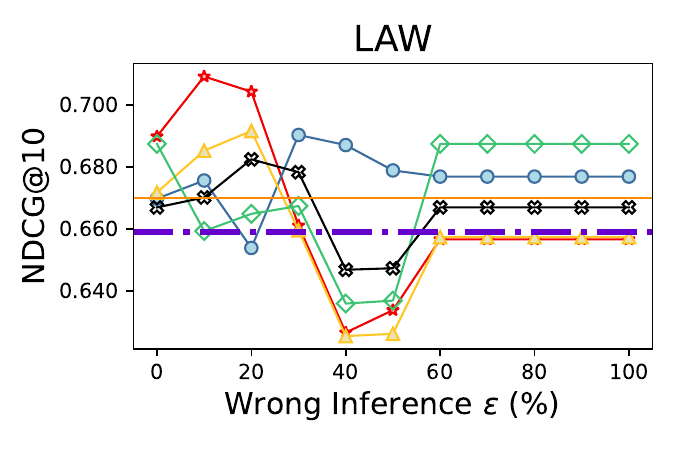}
    \end{subfigure}
    \hfill
    \begin{subfigure}{0.24\textwidth}
        \includegraphics[width=\linewidth]{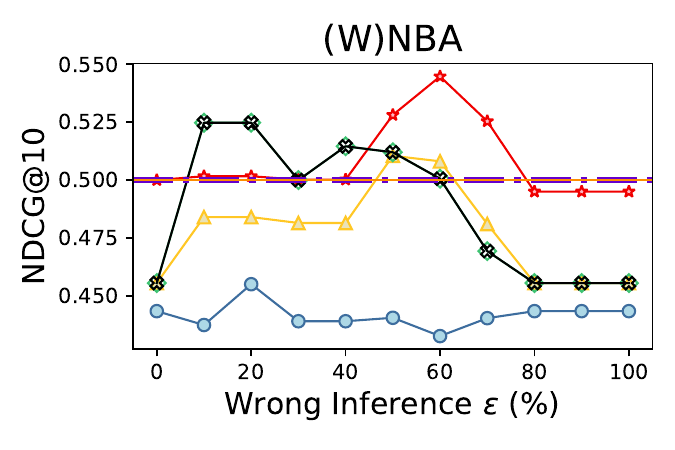}
    \end{subfigure}
    \hfill
    \begin{subfigure}{0.24\textwidth}
        \includegraphics[width=\linewidth]{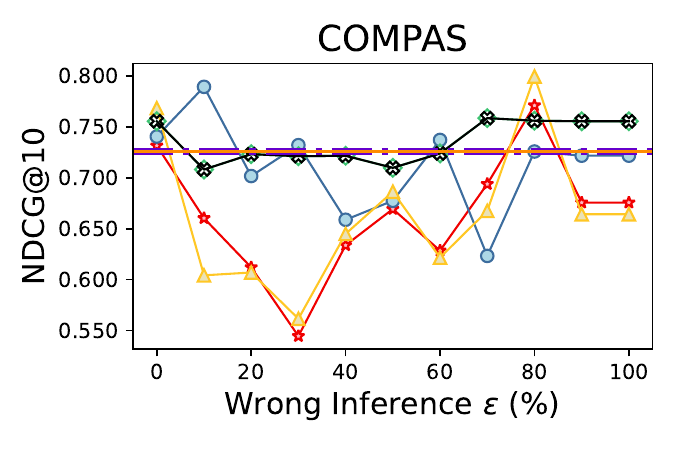}
    \end{subfigure}
    \hfill
    \begin{subfigure}{0.24\textwidth}
        \includegraphics[width=\linewidth]{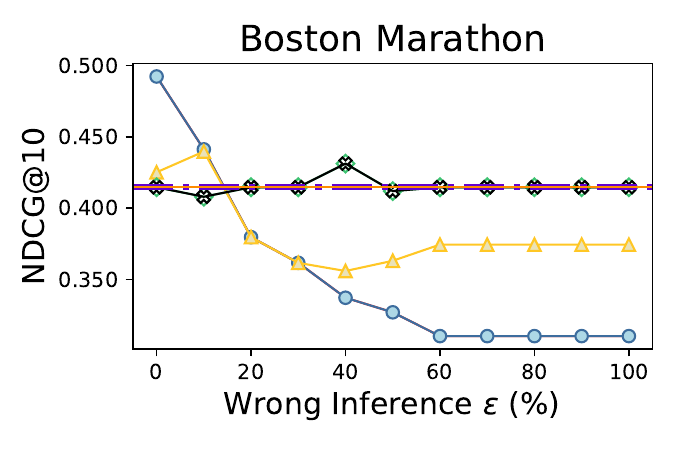}
    \end{subfigure}
    \hfill
    \begin{subfigure}{0.24\textwidth}
        \includegraphics[width=\linewidth]{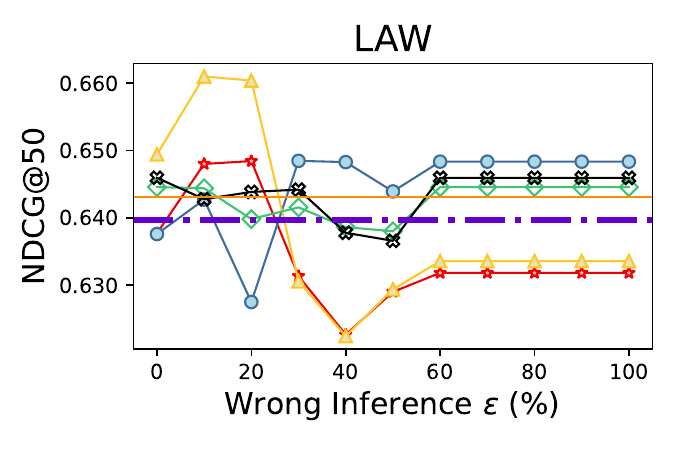}
    \end{subfigure}
    \hfill
    \begin{subfigure}{0.24\textwidth}
        \includegraphics[width=\linewidth]{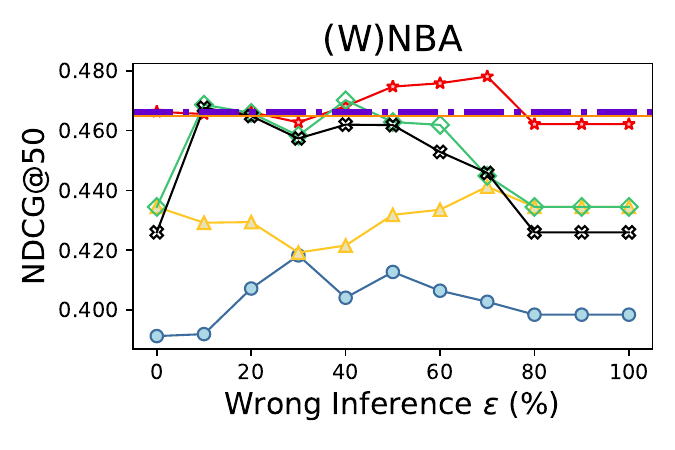}
        
    \end{subfigure}
    \hfill
    \begin{subfigure}{0.24\textwidth}
        \includegraphics[width=\linewidth]{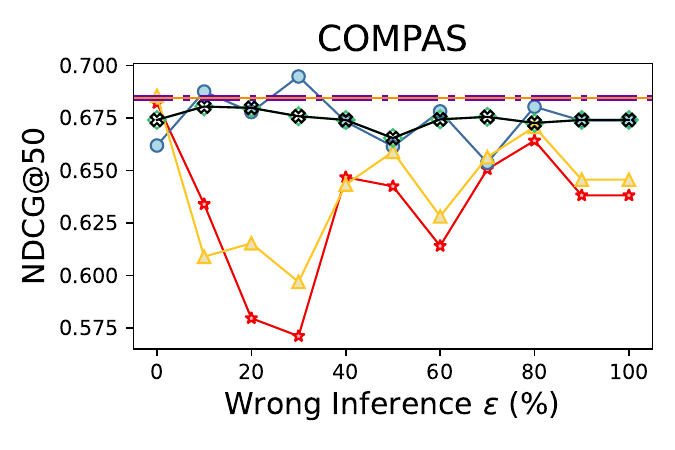}
    \end{subfigure}
    \hfill
    \begin{subfigure}{0.24\textwidth}
        {\includegraphics[width=\linewidth]{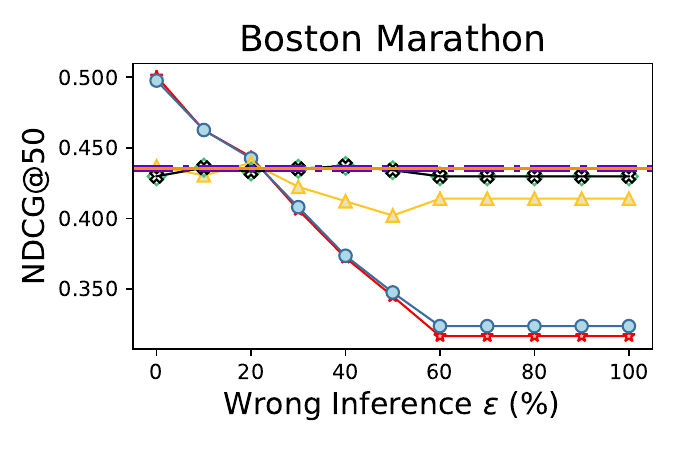}}
    \end{subfigure}   
    \hfill

    \caption{NDCG plots at different cutoffs ($k=10$ and $k=50$) for the 1$^\mathrm{st}$ simulation scenario  ($g_{dis}\leftrightarrow g_{adv}$).}
    \label{fig:extra1}
\end{figure*}

\begin{figure*}[ht]
\centering
\begin{subfigure}{1\textwidth}
\includegraphics[width=1\linewidth]{figures/CaseStudies/legend_case.pdf}
\end{subfigure}
\begin{subfigure}{0.3\linewidth}
\includegraphics[width=1\linewidth]{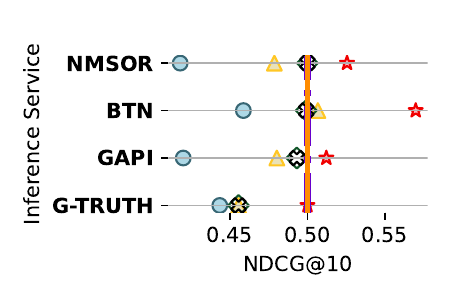}
\end{subfigure}
\begin{subfigure}{0.3\linewidth}
\includegraphics[width=1\linewidth]{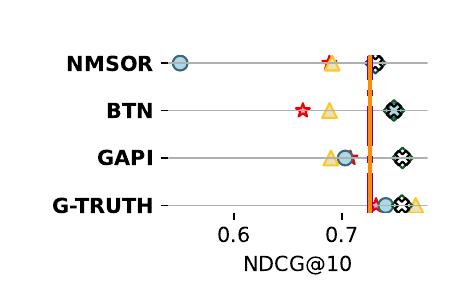}
\end{subfigure}
\begin{subfigure}{0.3\linewidth}
\includegraphics[width=1\linewidth]{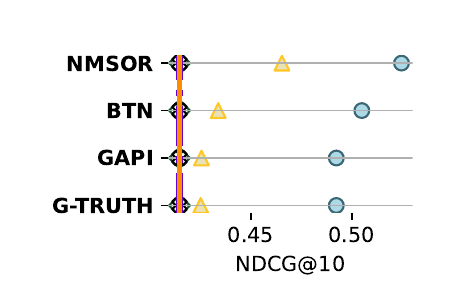}
\end{subfigure}
\begin{subfigure}{0.3\linewidth}
\includegraphics[width=1\linewidth]{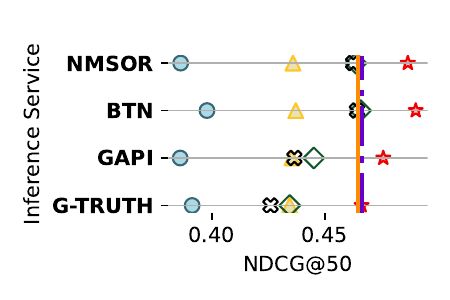}
\end{subfigure}
\begin{subfigure}{0.3\linewidth}
\includegraphics[width=1\linewidth]{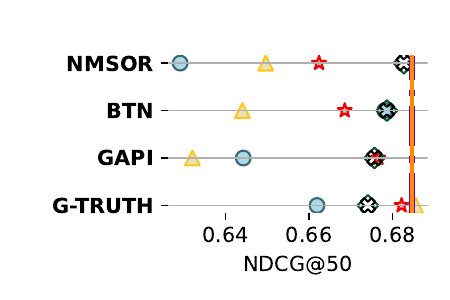}
\end{subfigure}
\begin{subfigure}{0.3\linewidth}
\includegraphics[width=1\linewidth]{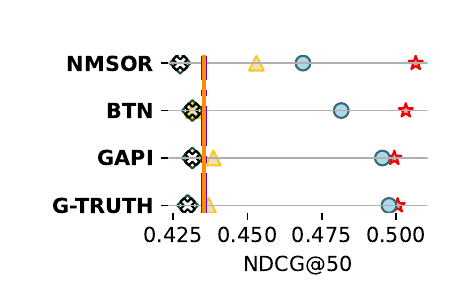}
\end{subfigure}

    \caption{NDCG plots at different cutoffs ($k=10$ and $k=50$) for real world use case experiments}
    \label{fig:extra2}

\end{figure*}

The NDCG@10 and NDCG@50 results for the bidirectional scenario ($g_{dis}\leftrightarrow g_{adv}$) and the case studies are displayed in Figures \ref{fig:extra1} and \ref{fig:extra2}. They are quite similar to the NDCG@100 in Section \ref{sect.results}. We however, see more consistent trends and a better interpretation of the fairness-utility trade-off in NDCG@100. This is because there are more candidate items that are representative of the underlying shift in exposure of the groups when there is error in inference. The NDCG@100 results for the real inference service experiments are similar to the results for the NDCG@10 and NDCG@50. 

\subsection{Unidirectional Results }
\label{sec:sim_oneand_two}
\begin{figure*}[ht]
    \centering

    \begin{subfigure}{1\textwidth}
        \includegraphics[width=\linewidth]{figures/Simulations/legend_synth.pdf}
        
    \end{subfigure}
    \hfill
    \begin{subfigure}{0.24\textwidth}    \includegraphics[width=\linewidth]{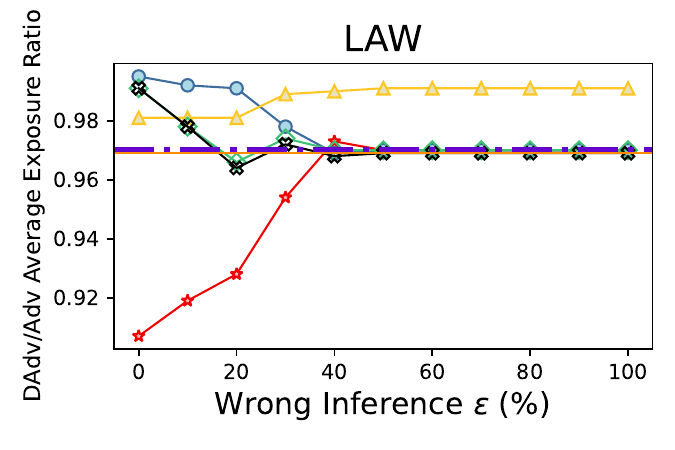}

    \end{subfigure}
    \hfill
    \begin{subfigure}{0.24\textwidth}
        \includegraphics[width=\linewidth]{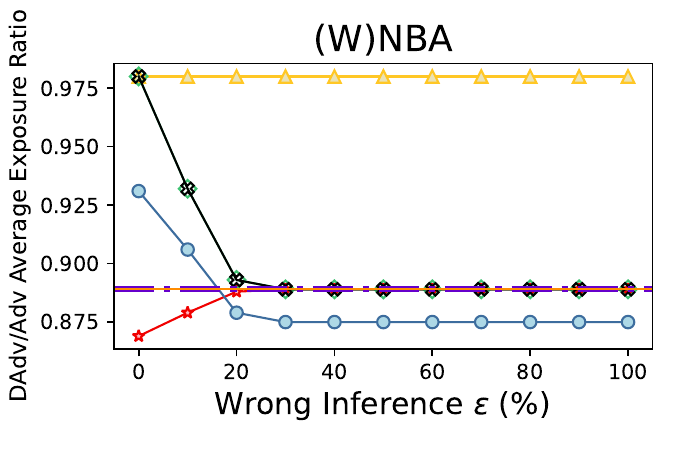}
    \end{subfigure}
    \hfill
    \begin{subfigure}{0.24\textwidth}
        \includegraphics[width=\linewidth]{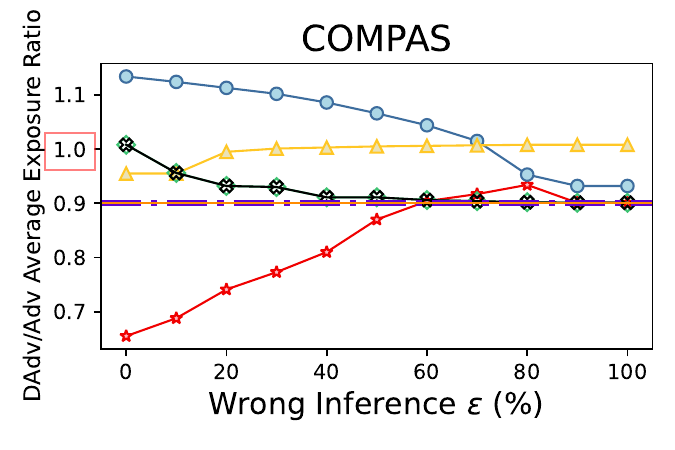}
    \end{subfigure}
    \hfill
    \begin{subfigure}{0.24\textwidth}
        \includegraphics[width=\linewidth]{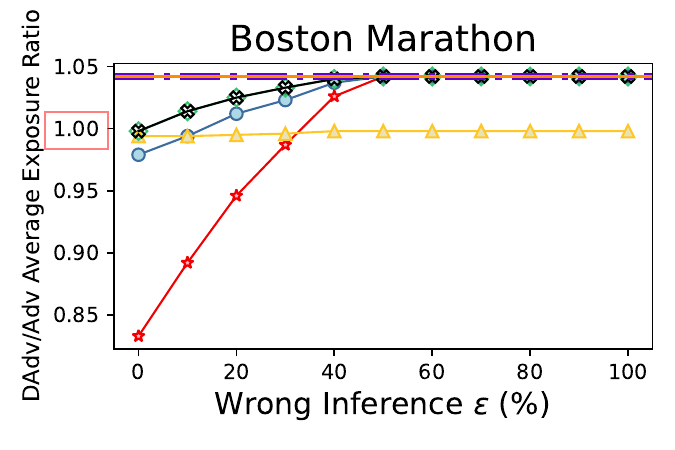}

    \end{subfigure}
    \hfill
    \begin{subfigure}{0.24\textwidth}
        \includegraphics[width=\linewidth]{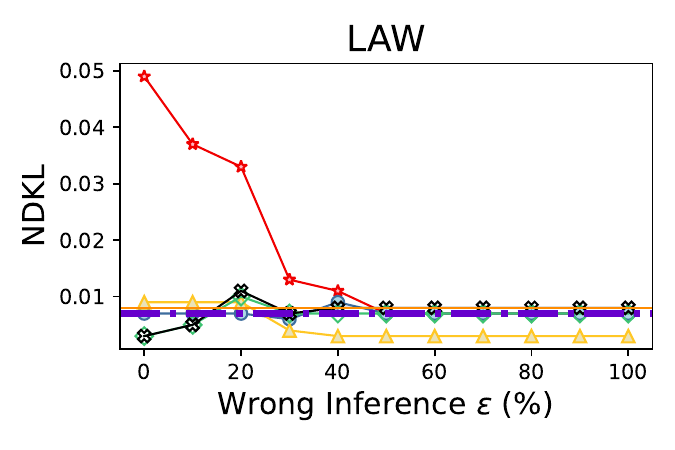}
    \end{subfigure}
    \hfill
    \begin{subfigure}{0.24\textwidth}
        \includegraphics[width=\linewidth]{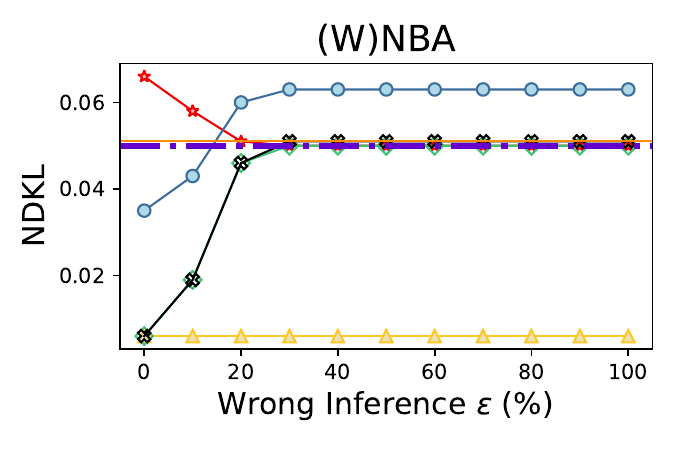}
        
    \end{subfigure}
    \hfill
    \begin{subfigure}{0.24\textwidth}
        \includegraphics[width=\linewidth]{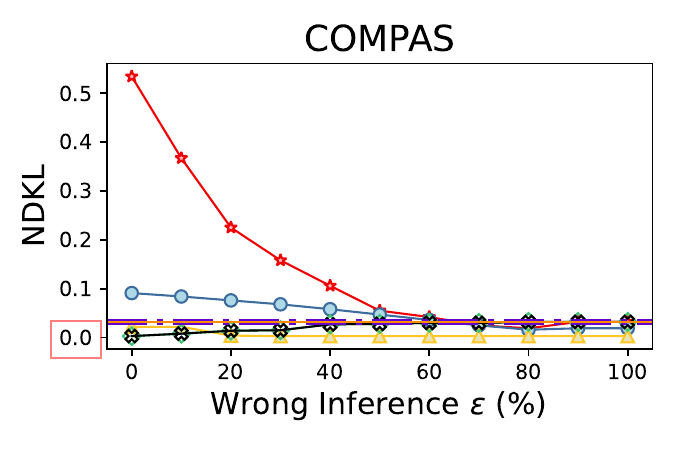}
        
    \end{subfigure}
    \hfill
    \begin{subfigure}{0.24\textwidth}
        {\includegraphics[width=\linewidth]{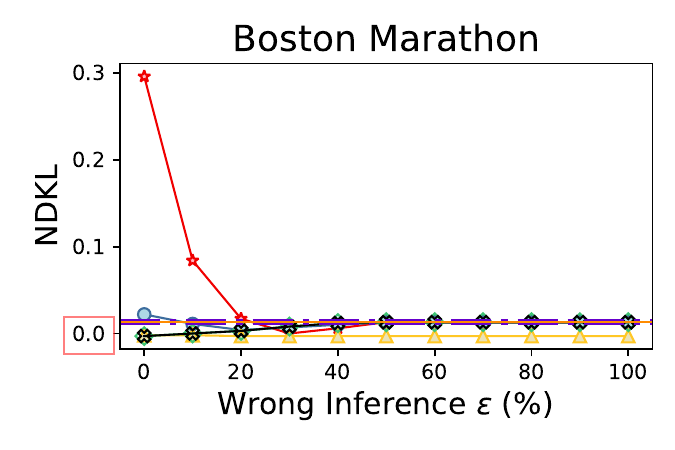}}
    \end{subfigure}   
    \hfill
     \begin{subfigure}{0.24\textwidth}
        \includegraphics[width=\linewidth]{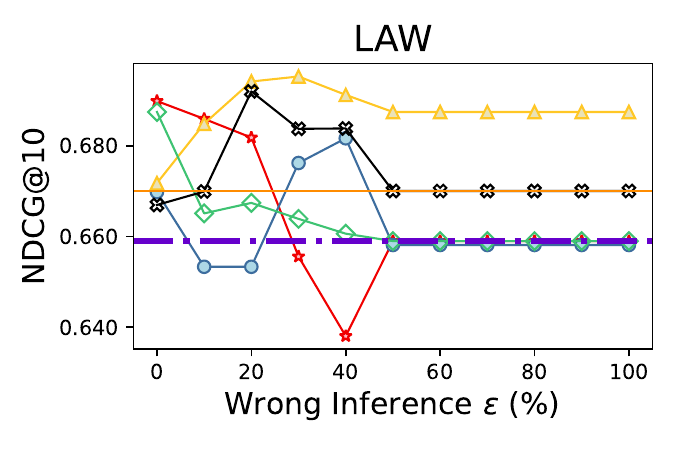}
    \end{subfigure}
    \hfill
    \begin{subfigure}{0.24\textwidth}
        \includegraphics[width=\linewidth]{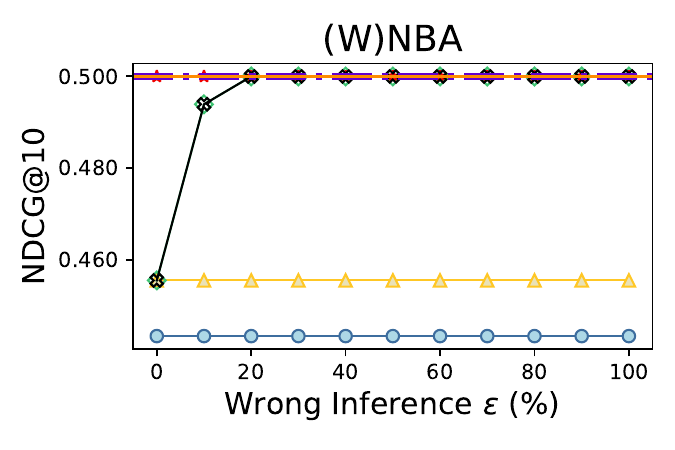}
        
    \end{subfigure}
    \hfill
    \begin{subfigure}{0.24\textwidth}
        \includegraphics[width=\linewidth]{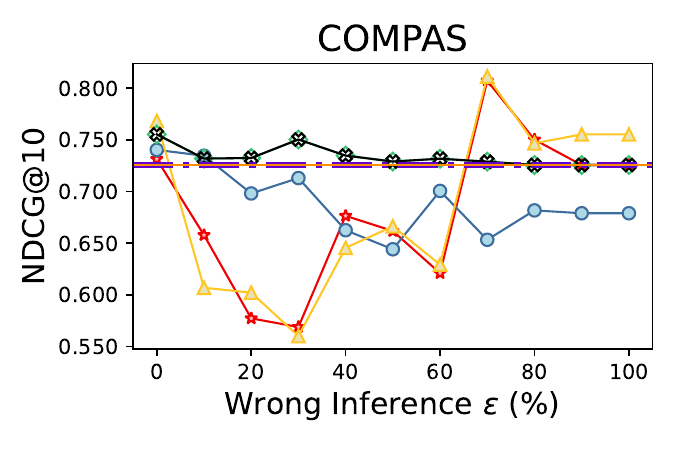}
    \end{subfigure}
    \hfill
    \begin{subfigure}{0.24\textwidth}
        \includegraphics[width=\linewidth]{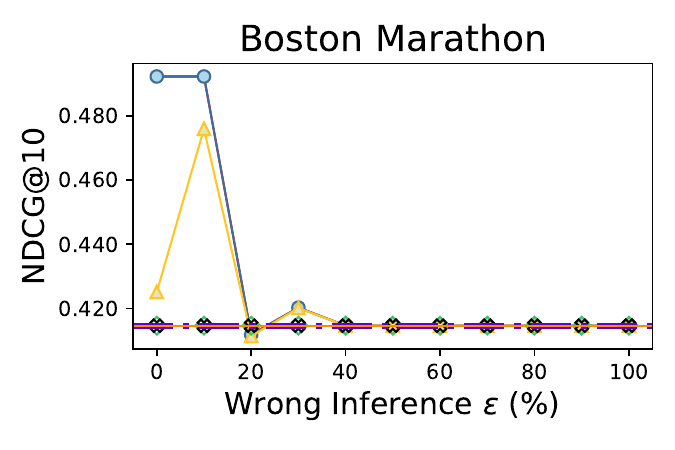}
    \end{subfigure}
     \begin{subfigure}{0.24\textwidth}
        \includegraphics[width=\linewidth]{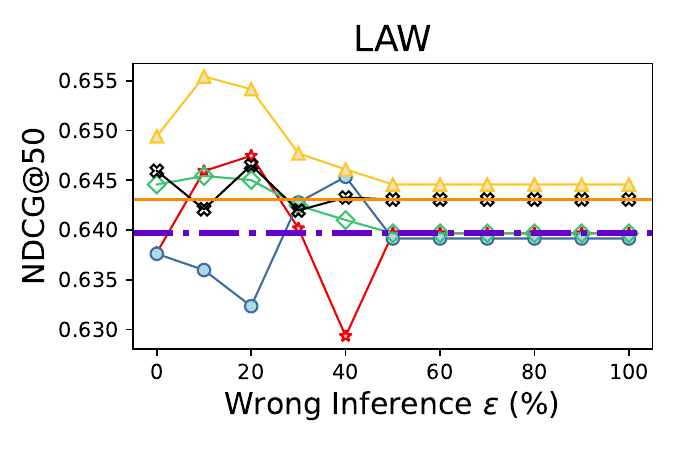}
    \end{subfigure}
    \hfill
    \begin{subfigure}{0.24\textwidth}
        \includegraphics[width=\linewidth]{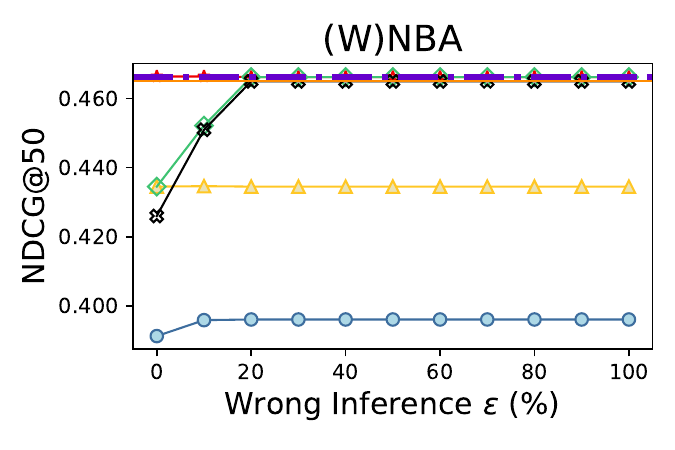}
        
    \end{subfigure}
    \hfill
    \begin{subfigure}{0.24\textwidth}
        \includegraphics[width=\linewidth]{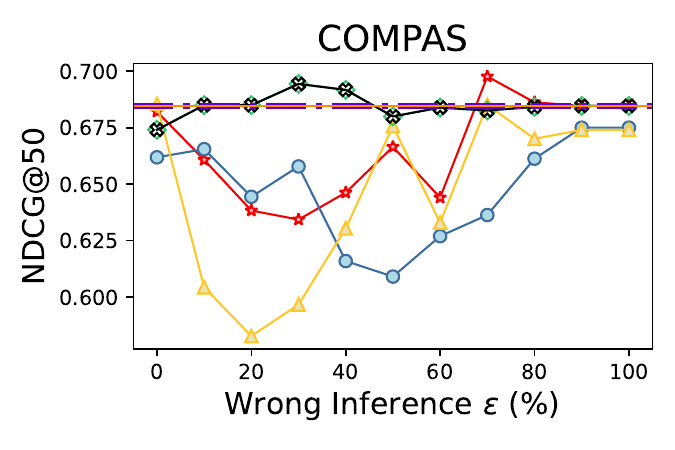}
    \end{subfigure}
    \hfill
    \begin{subfigure}{0.24\textwidth}
        \includegraphics[width=\linewidth]{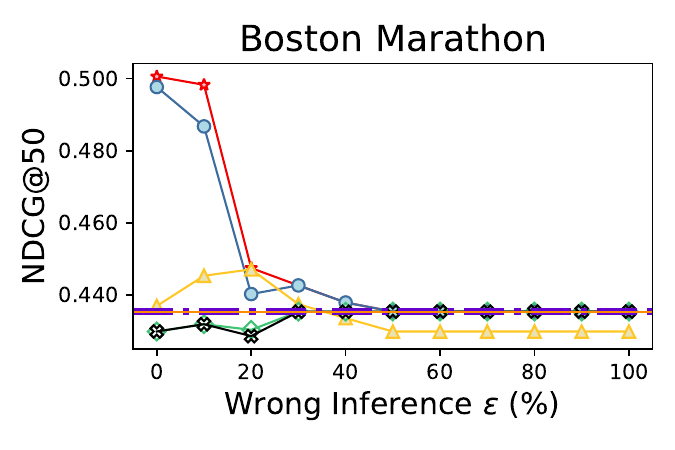}
    \end{subfigure}
    
    \begin{subfigure}{0.24\textwidth}
        \includegraphics[width=\linewidth]{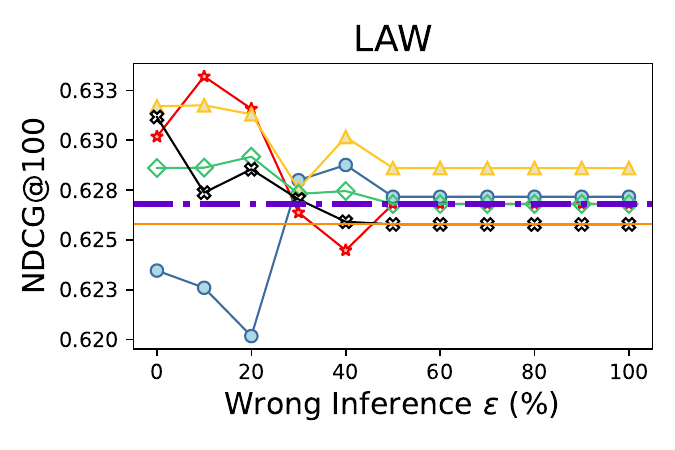}
    \end{subfigure}
    \hfill
    \begin{subfigure}{0.24\textwidth}
        \includegraphics[width=\linewidth]{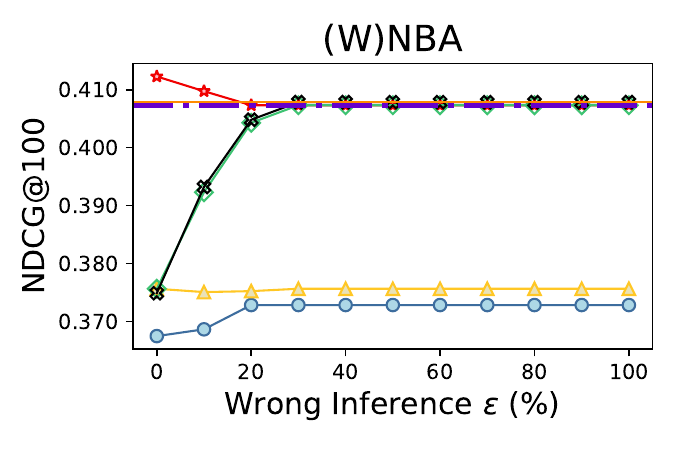}
        
    \end{subfigure}
    \hfill
    \begin{subfigure}{0.24\textwidth}
        \includegraphics[width=\linewidth]{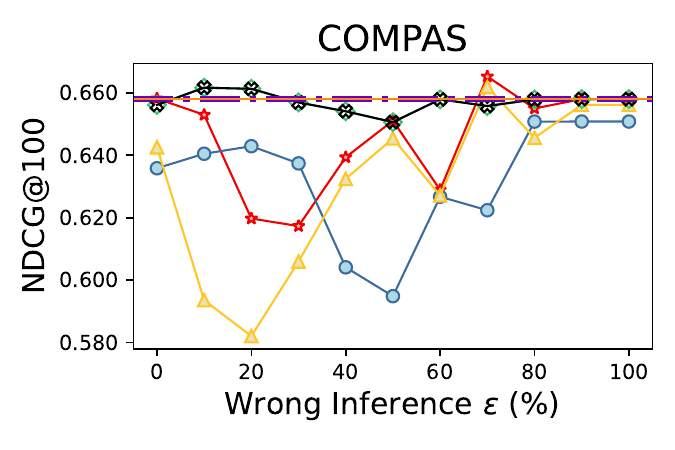}
    \end{subfigure}
    \hfill
    \begin{subfigure}{0.24\textwidth}
        \includegraphics[width=\linewidth]{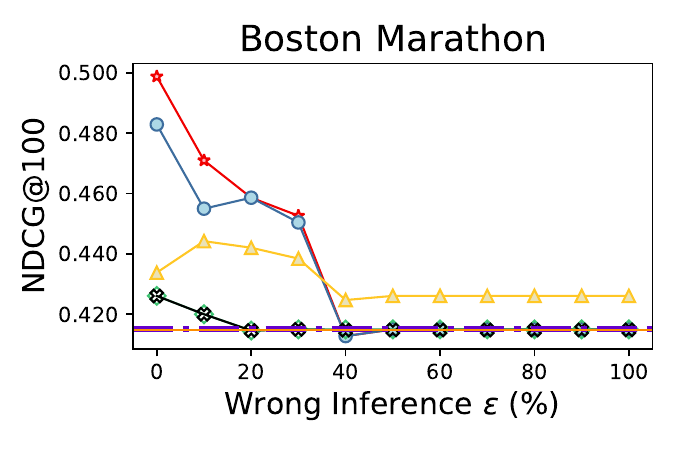}
    \end{subfigure}
    \caption{DAdv/Adv Exposure Ratio, NDKL and NDCG graphs for the 2$^\mathrm{nd}$ simulation scenario and each model, $g_{dis}\rightarrow g_{adv}$}
    \label{fig:sims2}
\end{figure*}

\begin{figure*}[ht]
    \centering

    \begin{subfigure}{1\textwidth}
        \includegraphics[width=\linewidth]{figures/Simulations/legend_synth.pdf}
        
    \end{subfigure}
    \hfill
    \begin{subfigure}{0.24\textwidth}    \includegraphics[width=\linewidth]{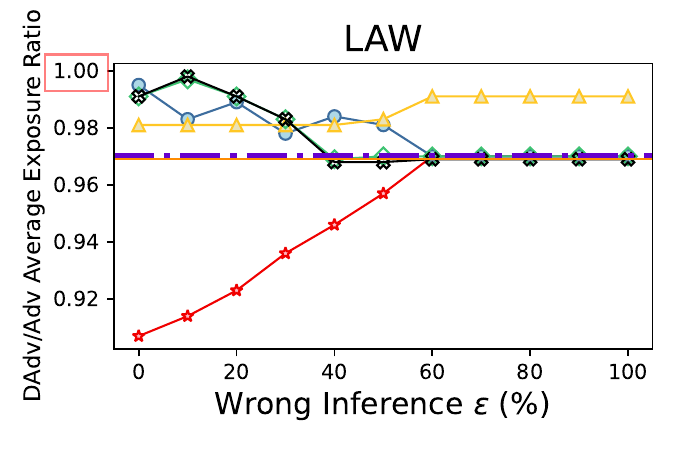}

    \end{subfigure}
    \hfill
    \begin{subfigure}{0.24\textwidth}
        \includegraphics[width=\linewidth]{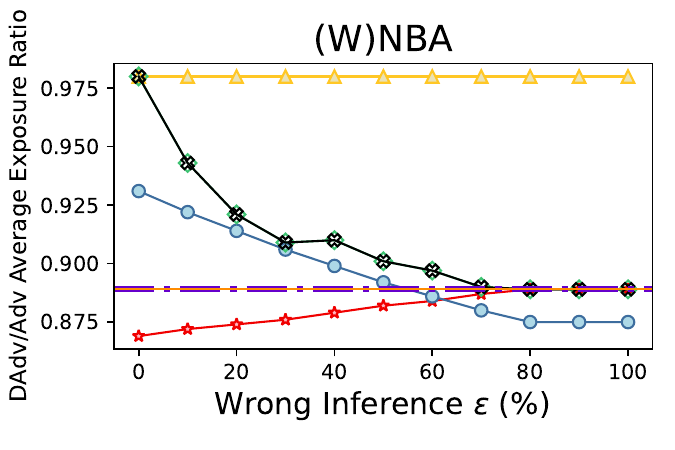}
    \end{subfigure}
    \hfill
    \begin{subfigure}{0.24\textwidth}
        \includegraphics[width=\linewidth]{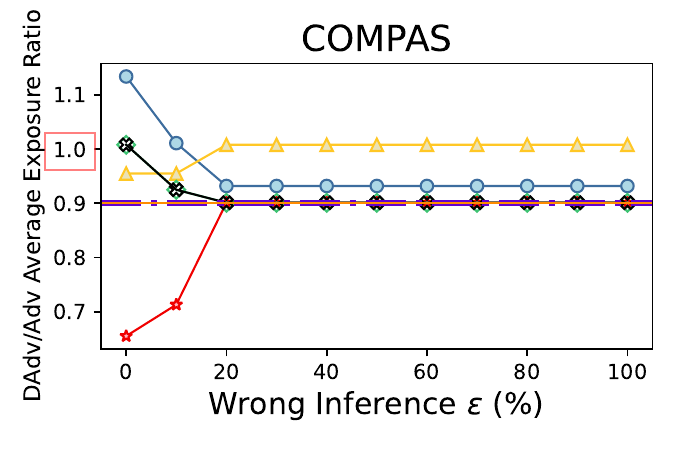}
    \end{subfigure}
    \hfill
    \begin{subfigure}{0.24\textwidth}
        \includegraphics[width=\linewidth]{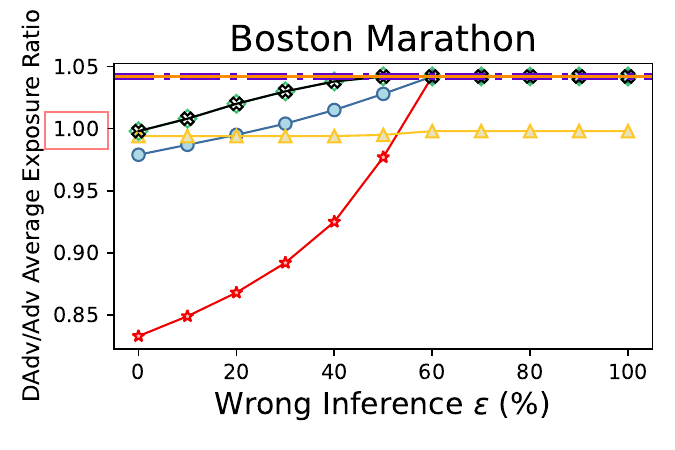}

    \end{subfigure}
    \hfill
    \begin{subfigure}{0.24\textwidth}
        \includegraphics[width=\linewidth]{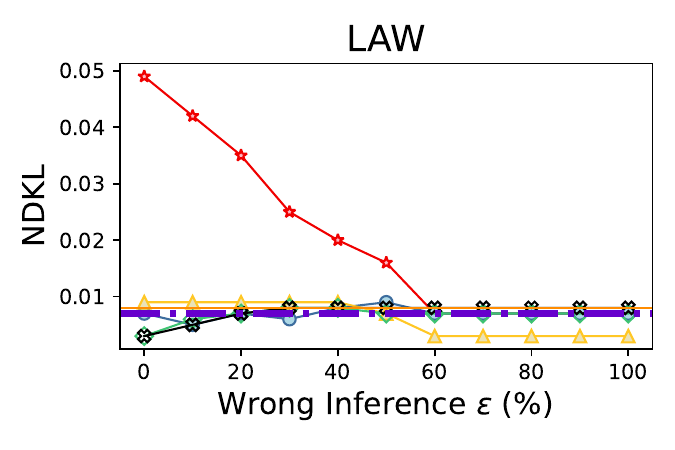}
    \end{subfigure}
    \hfill
    \begin{subfigure}{0.24\textwidth}
        \includegraphics[width=\linewidth]{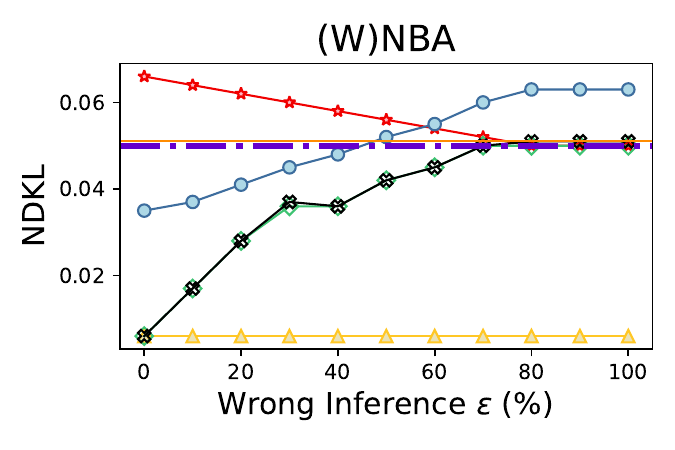}
        
    \end{subfigure}
    \hfill
    \begin{subfigure}{0.24\textwidth}
        \includegraphics[width=\linewidth]{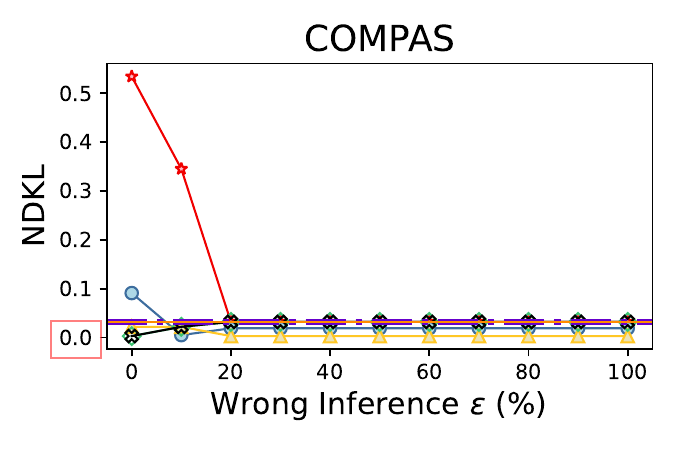}
        
    \end{subfigure}
    \hfill
    \begin{subfigure}{0.24\textwidth}
        {\includegraphics[width=\linewidth]{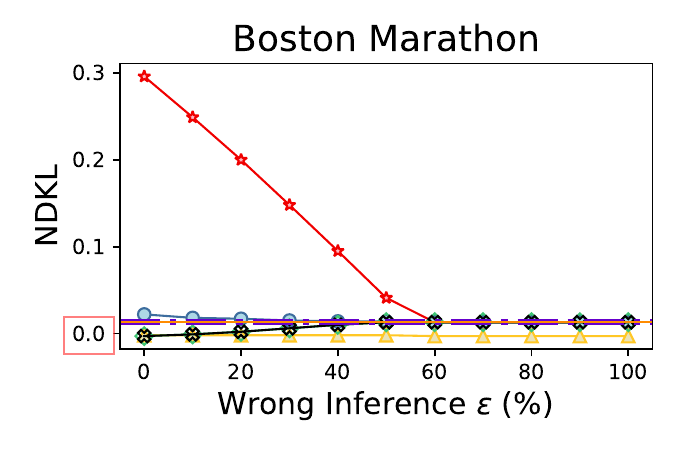}}
    \end{subfigure}   

    \begin{subfigure}{0.24\textwidth}
        \includegraphics[width=\linewidth]{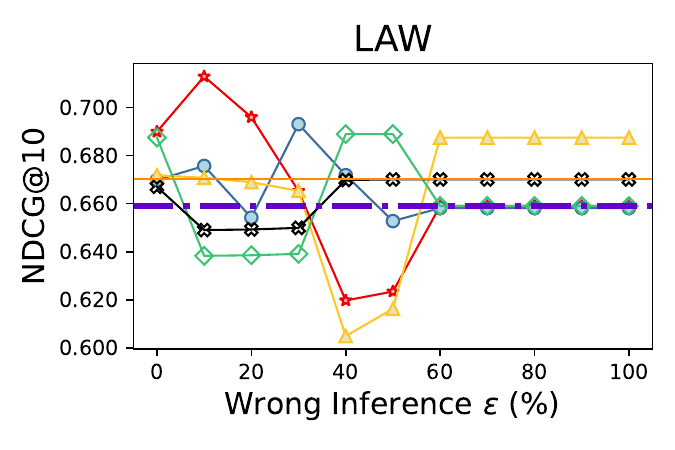}
    \end{subfigure}
    \hfill
    \begin{subfigure}{0.24\textwidth}
        \includegraphics[width=\linewidth]{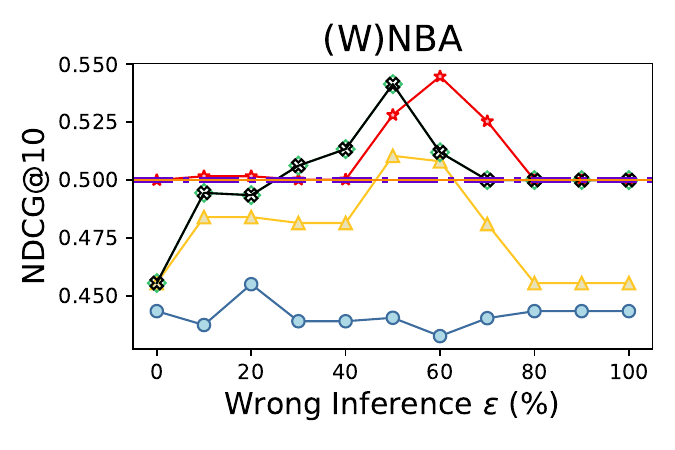}
        
    \end{subfigure}
    \hfill
    \begin{subfigure}{0.24\textwidth}
        \includegraphics[width=\linewidth]{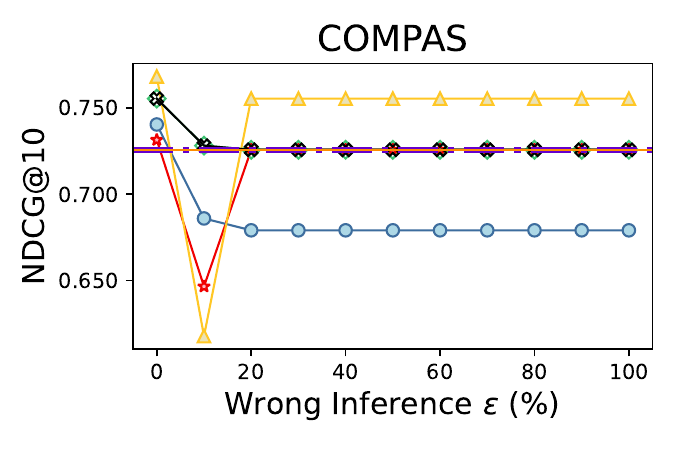}
    \end{subfigure}
    \hfill
    \begin{subfigure}{0.24\textwidth}
        \includegraphics[width=\linewidth]{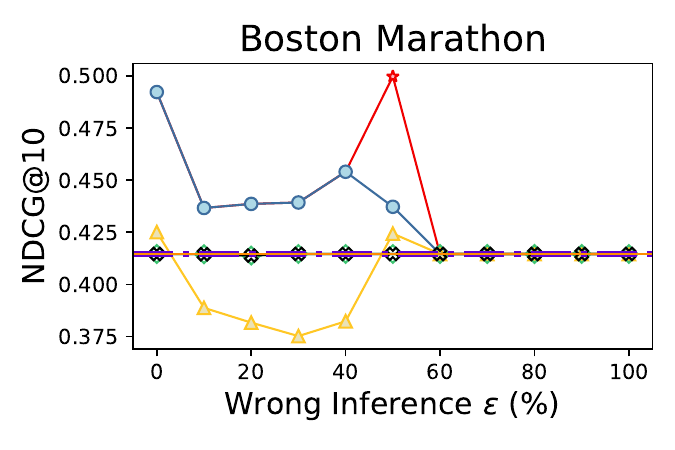}
    \end{subfigure}
    \begin{subfigure}{0.24\textwidth}
        \includegraphics[width=\linewidth]{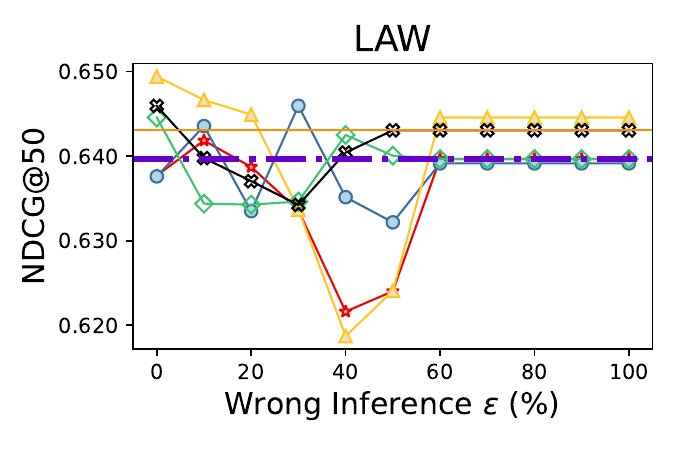}
    \end{subfigure}
    \hfill
    \begin{subfigure}{0.24\textwidth}
        \includegraphics[width=\linewidth]{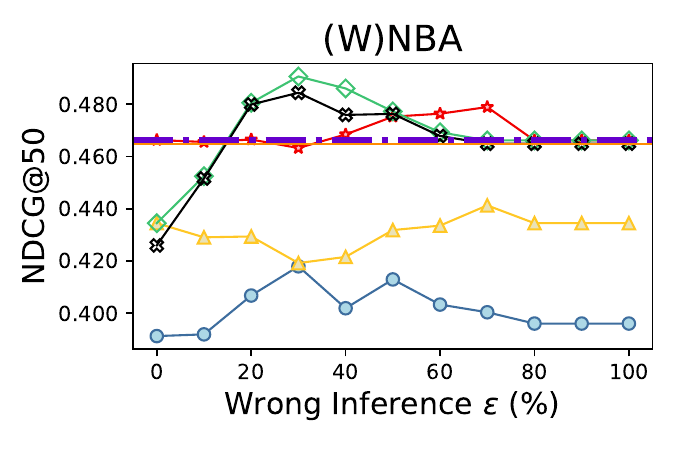}
        
    \end{subfigure}
    \hfill
    \begin{subfigure}{0.24\textwidth}
        \includegraphics[width=\linewidth]{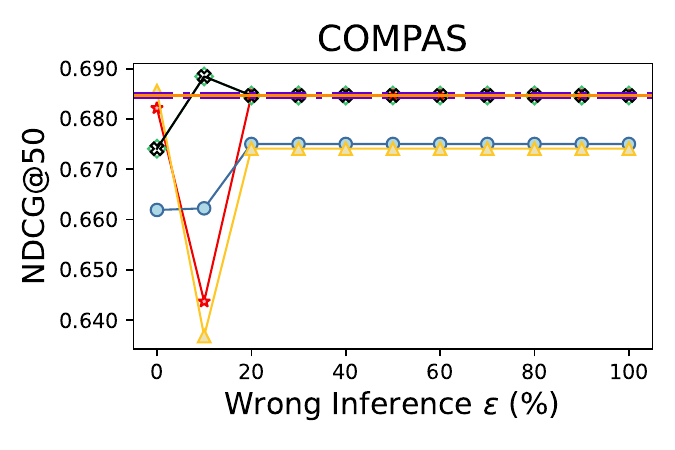}
    \end{subfigure}
    \hfill
    \begin{subfigure}{0.24\textwidth}
        \includegraphics[width=\linewidth]{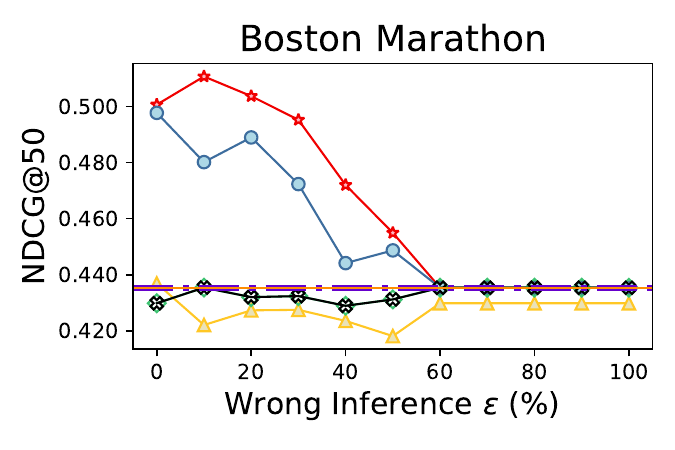}
    \end{subfigure}
    \hfill
    
    \begin{subfigure}{0.24\textwidth}
        \includegraphics[width=\linewidth]{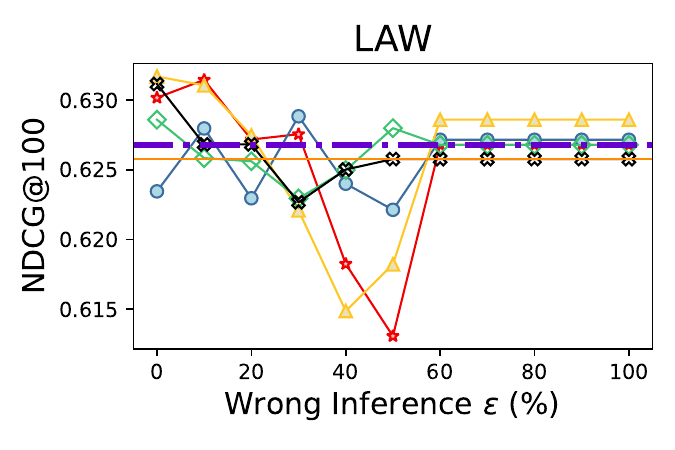}
    \end{subfigure}
    \hfill
    \begin{subfigure}{0.24\textwidth}
        \includegraphics[width=\linewidth]{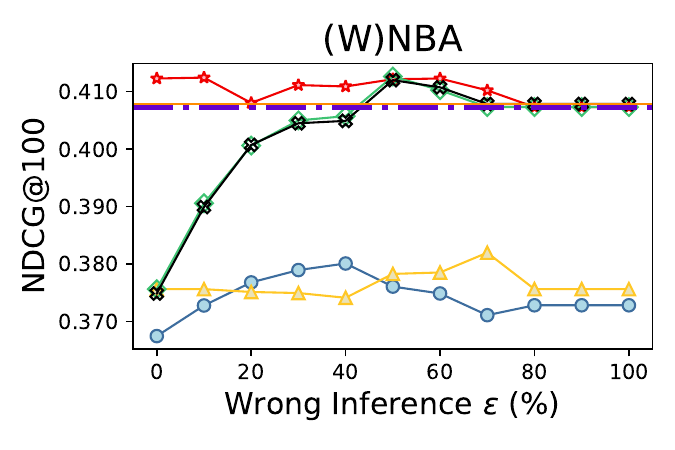}
        
    \end{subfigure}
    \hfill
    \begin{subfigure}{0.24\textwidth}
        \includegraphics[width=\linewidth]{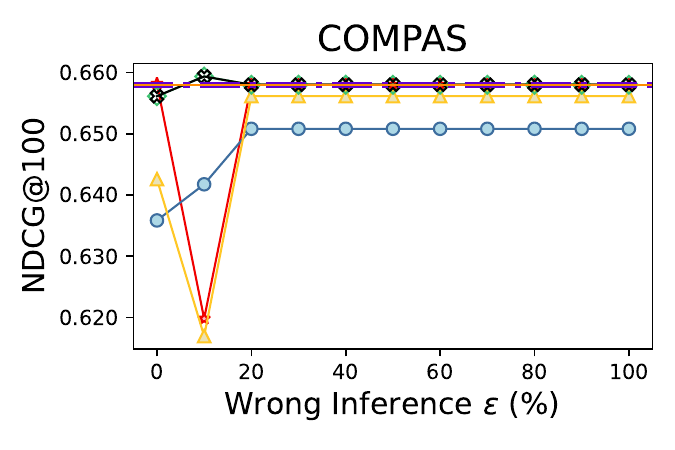}
    \end{subfigure}
    \hfill
    \begin{subfigure}{0.24\textwidth}
        \includegraphics[width=\linewidth]{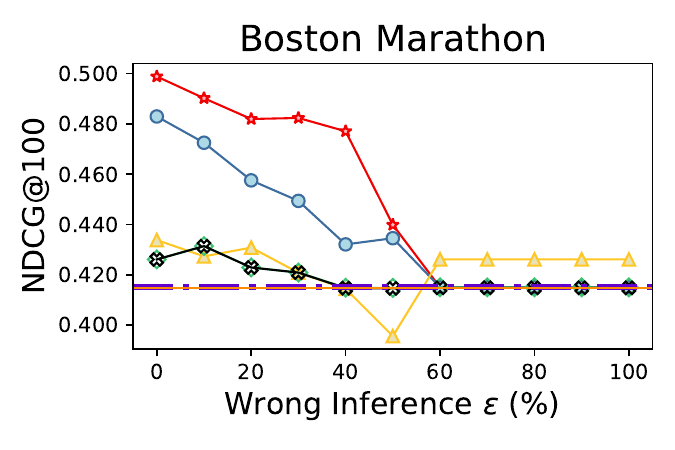}
    \end{subfigure}
    \caption{DAdv/Adv Exposure Ratio, NDKL and NDCG graphs for the 3$^\mathrm{rd}$ simulation scenario and each model, $g_{dis}\leftarrow g_{adv}$}
    \label{fig:sims3}
\end{figure*}

The \dadexp{}, NDKL and NDCG graphs for the 2$^\mathrm{nd}$  and 3$^\mathrm{rd}$ simulation scenarios are shown in Figures \ref{fig:sims2} and \ref{fig:sims3}.

\begin{description}
 \item [\algoone{}] \ 
\begin{itemize}[leftmargin=1pt, align=left, labelindent=0pt]
\item Similar results as in the 1$^\mathrm{st}$ simulation scenario.
\end{itemize}

 \item [\algotwo{}] \ 
\begin{itemize}[leftmargin=1pt, align=left, labelindent=0pt]
\item For \algotwo, the \dadexp{} never go beyond \algoone{} and \algothree{} across all inference error rates, converging to these values. 
\item The NDKL values also converge to the NDKL for \algoone{} and \algothree{}. 
\item The NDCG values show a similar behaviour to the 1$^\mathrm{st}$ simulation scenario.
\end{itemize}

 \item [\algothree{}] \ 
\begin{itemize}[leftmargin=1pt, align=left, labelindent=0pt]
\item Similar results as in the 1$^\mathrm{st}$ simulation scenario.
\end{itemize}

 \item [\algofour{}] \ 
\begin{itemize}[leftmargin=1pt, align=left, labelindent=0pt]
\item We get similar \dadexp{} results as in the 1$^\mathrm{st}$ simulation scenario.
\item For the NDKL, the results are similar only for (W)NBA dataset. 
\item We get similar NDCG results as in the 1$^\mathrm{st}$ simulation scenario.
\end{itemize}

\item [\algofive{}] \ 
\begin{itemize}[leftmargin=1pt, align=left, labelindent=0pt]

\item We see that the \dadexp{} for $0$ and $100 \%$ are not the same as in 1$^\mathrm{st}$ simulation scenario, since we do not have the same complete swap situation here.

\item Similar NDKL results as in the 1$^\mathrm{st}$ simulation scenario.

\item  For (W)NBA, NDCG shows a sharp growth until $\epsilon=50\%$, until it reaches a similar value to \algothree{}.

\end{itemize} 

 \item [\algosix{}] \ 
\begin{itemize}[leftmargin=1pt, align=left, labelindent=0pt]
\item Similar results as in the 1$^\mathrm{st}$ simulation scenario.
\end{itemize}

 \item [\algoseven{}] \ 
\begin{itemize}[leftmargin=1pt, align=left, labelindent=0pt]
\item Similar results as in the 1$^\mathrm{st}$ simulation scenario. 
\end{itemize}
\end{description}

\subsection{Model Architecture, Training and  Parameter Settings} \label{sec:training_2}

\subsubsection{Model Architecture}

For each strategy, there is a training and testing (ranking) stage, with an optional re-ranking stage (See Table \ref{tab:overview_strategies}).

\begin{figure}[ht]
\begin{subfigure}{.5\textwidth}
\includegraphics[width=.8\linewidth]{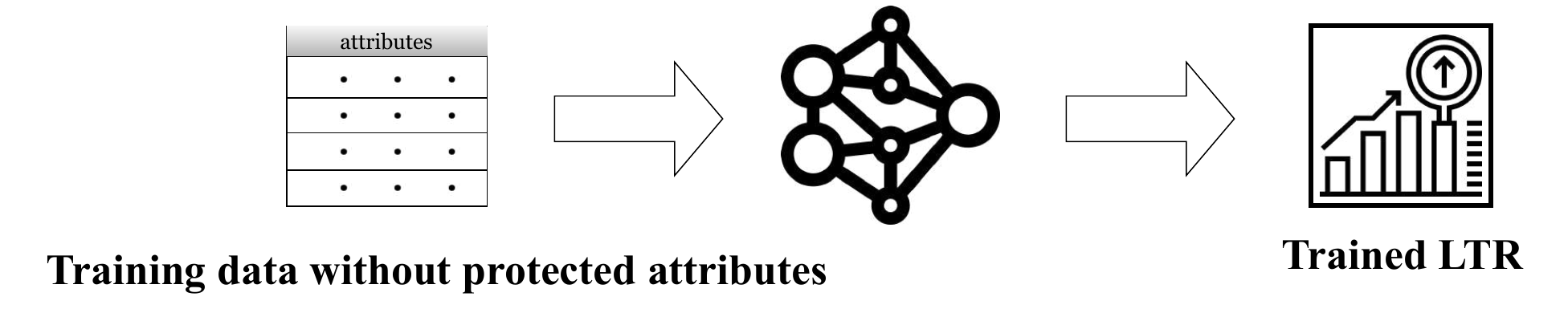}
\caption*{Training Stage without protected attributes  - LTR model}
\end{subfigure}
\begin{subfigure}{.5\textwidth}

\includegraphics[width=.75\linewidth]{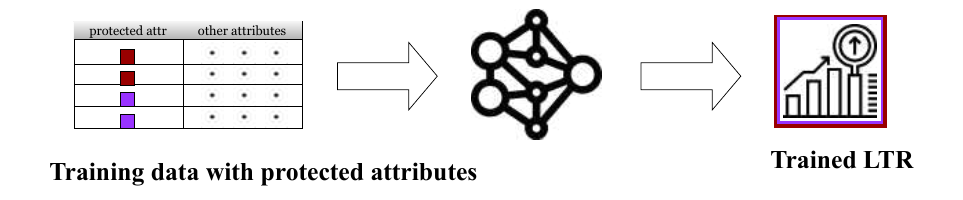}
\caption*{Training Stage with protected attributes  - LTR model}
 \end{subfigure} \begin{subfigure}{.5\textwidth}
\includegraphics[width=.9\linewidth]{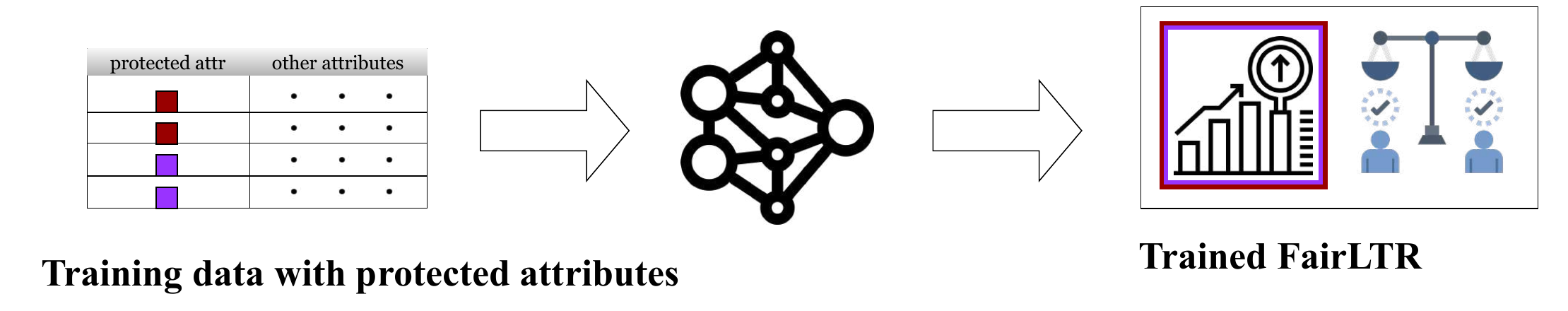}
\caption*{Training Stage with protected attributes  - FairLTR model}
\end{subfigure} 

\caption{Training Architecture}
    \label{fig:architecture-training}
\end{figure}

\subsubsection{Training.}

For each dataset in Section \ref{sec:datasets}, three learning-to-rank models are trained: a fairness-unaware model
trained 
without the protected attributes, a fairness-unaware model trained reliant on the protected attributes, and a fairness-aware model requiring the protected attributes (Figure \ref{fig:architecture-training}).
 Each model takes list of candidate items,
each item having corresponding 
attributes
and a target score (judgment). The model learns a scoring function based on these attributes (scores and features). While the fairness-unware models ($\gamma = 0$) aims to reduce ranking errors, the third model ($\gamma > 0$) works to reduce both the ranking error and simultaneously the  \textit{disparate exposure} between two groups.
\begin{description}[wide,labelindent=0pt,parsep=0pt,itemsep=0pt,style=unboxed]
    \item[Listnet trained w/o protected attributes.] To train our Listnet model without protected attributes, we train  our model by setting gamma to zero and setting 
    all protected attribute values to 1 (yields the same results as setting to 0).
    \item[Listnet trained with protected attributes.] To train our Listnet model without protected attributes, we train  our model with ground-truth protected attributes, setting gamma to zero.
    \item[DELTR.] To train a fair DELTR model, gamma has to be greater than 1. To obtain a suitable gamma value, we either set the value to $L/U$ as described in \cite{DELTR} or by trial and error (for the Boston Marathon dataset), set the value high enough and plot loss graphs to ensure convergence. Note that DELTR works such that once the fairness goal is achieved (i.e., no disparate exposure), higher values of gamma will not affect fairness.
    
    \item[DetConstSort.] The inputs to the DetConstSort algorithm were as follows:
    \begin{enumerate}
        \item $a$: A list of the protected attributes 
        for the candidate items. Each item belongs either to the protected or non-protected group
        \item $k_{\max}$: This specifies the length of the ranking that you require the algorithm to return. For our experiments, we required the algorithm to return the entire list as a reranked list, therefore, $k_{\max}$ was the same as the length of the input ranking. 
        \item $p$: The DetConstSort algorithm  requires that a desired  categorical distribution of each group is specified. For our experiments, we set this to the underlying distribution of the list to be ranked.
    \end{enumerate}
\end{description}
\subsubsection{Testing (Ranking) Stages.}
\begin{figure*}

  \begin{subfigure}{.5\textwidth}

\includegraphics[width=.9\linewidth]{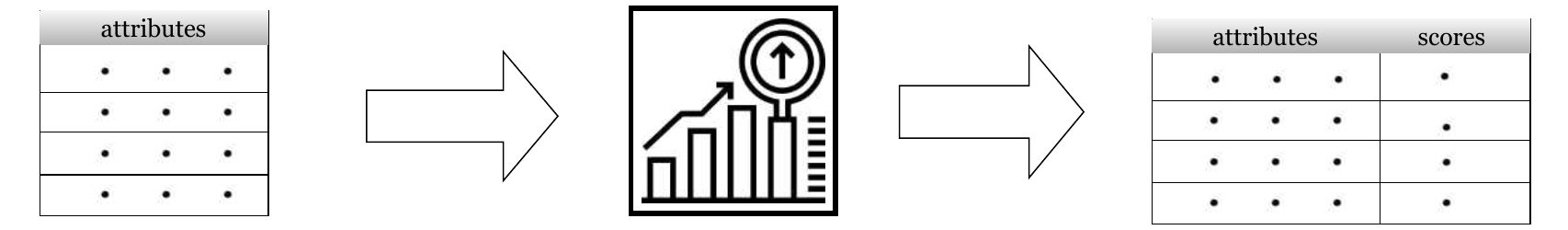}

\caption*{\algoone{}}
\end{subfigure}
\begin{subfigure}{.5\textwidth}

\includegraphics[width=1\linewidth]{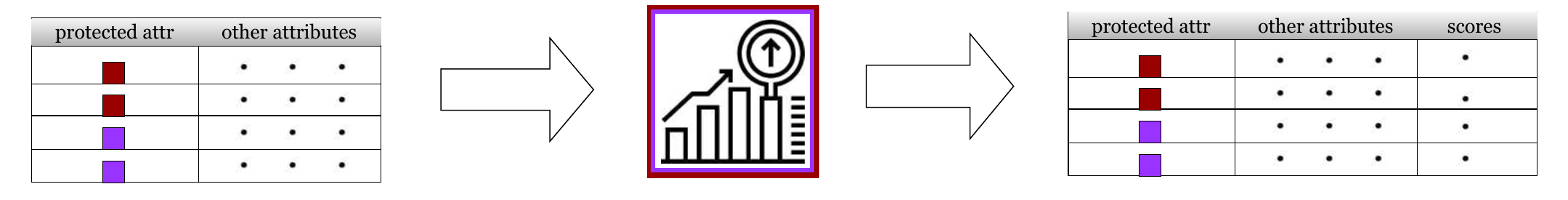}
\caption*{\algotwo{}}
\end{subfigure} \begin{subfigure}{.5\textwidth}
\includegraphics[width=1\linewidth]{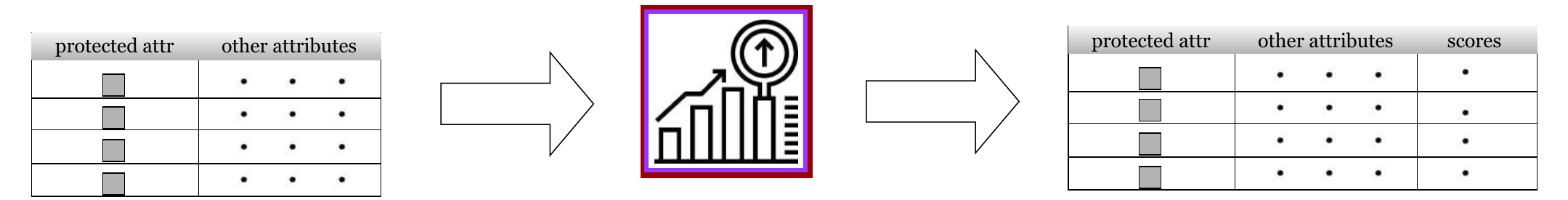}
\caption*{\algothree{}}
\end{subfigure}  
\begin{subfigure}{.5\textwidth}
\includegraphics[width=1\linewidth]{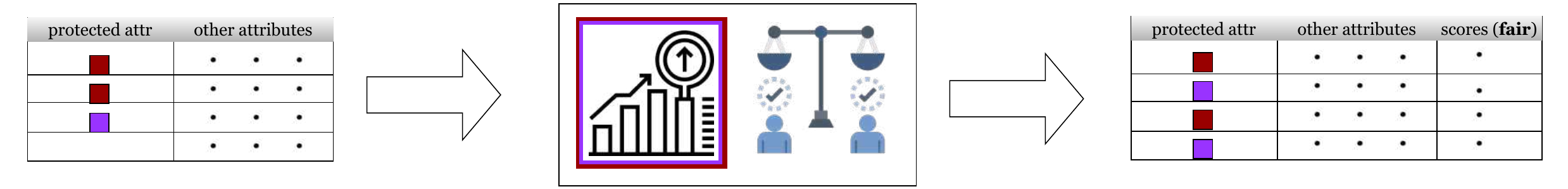}
\caption*{\algofour{}}
\end{subfigure}
\begin{subfigure}{1\textwidth}
\includegraphics[width=1\linewidth]{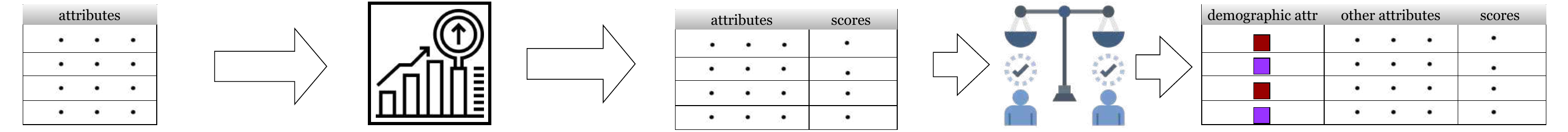}
\caption*{\algofive{}}
\end{subfigure}
\begin{subfigure}{1\textwidth}
\includegraphics[width=1\linewidth]{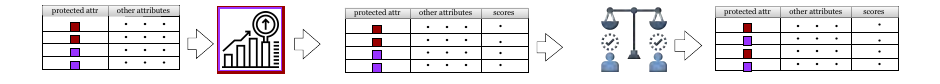}
\caption*{\algosix{}}
\end{subfigure}
\begin{subfigure}{1\textwidth}
\includegraphics[width=1\linewidth]{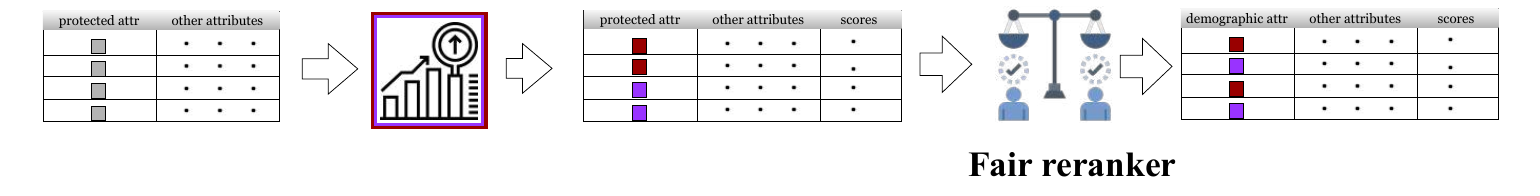}
\caption*{\algoseven{}}
\end{subfigure}
\caption{Ranking Strategies}
\label{fig:architecture-ranking}
\end{figure*}

Figure \ref{fig:architecture-ranking} shows model architecture for the ranking strategies.

\subsection{Parameters for DELTR and DetConstSort}
For DELTR, we started with values for the parameter $\gamma$ based on the process described in the related literature.
We verified that  the model training was converging for each of our data sets. As needed, we then
experimented with additional values doubling the values to verify the stability of our results and
convergence of the training. 

For DetConstSort, we set the target distribution to match the proportion of candidate items from each group in the ranking set, hence optimizing group fairness.

\label{sec-synthetic-study}

\subsection{Trade-Off Graphs for (W)NBA (NDKL vs NDCG)}
\begin{figure*}[ht]
    \centering
    \begin{subfigure}{\textwidth}    \includegraphics[width=1\linewidth]{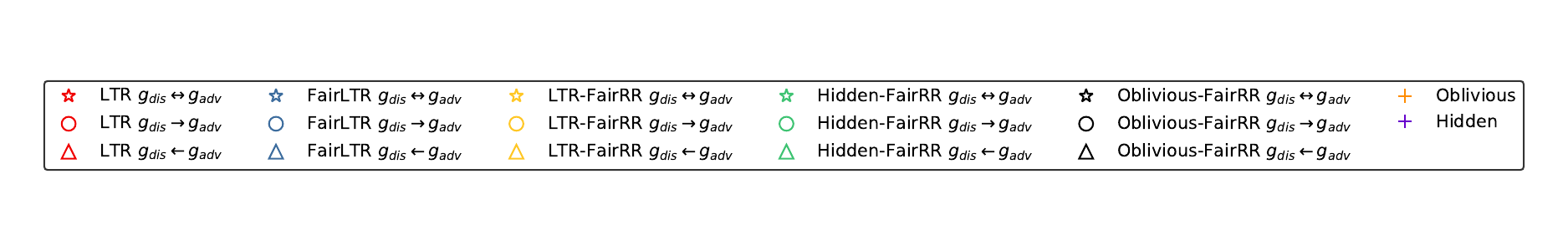}
        \end{subfigure}
    \begin{subfigure}{0.2\textwidth}    \includegraphics[width=\linewidth]{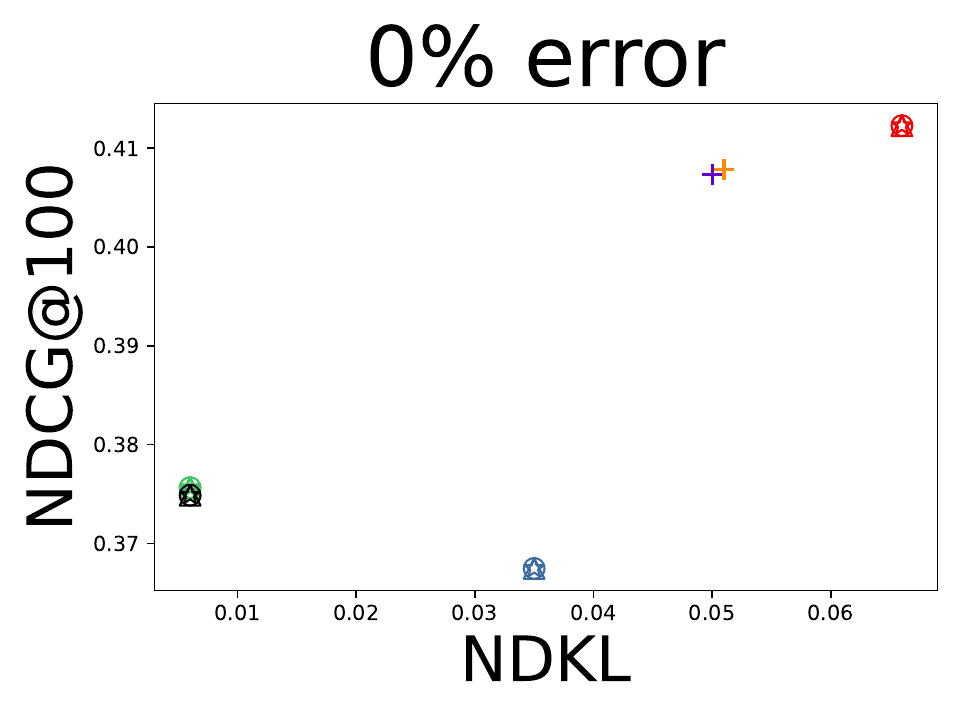}
    \end{subfigure}
    \begin{subfigure}{0.2\textwidth}    \includegraphics[width=\linewidth]{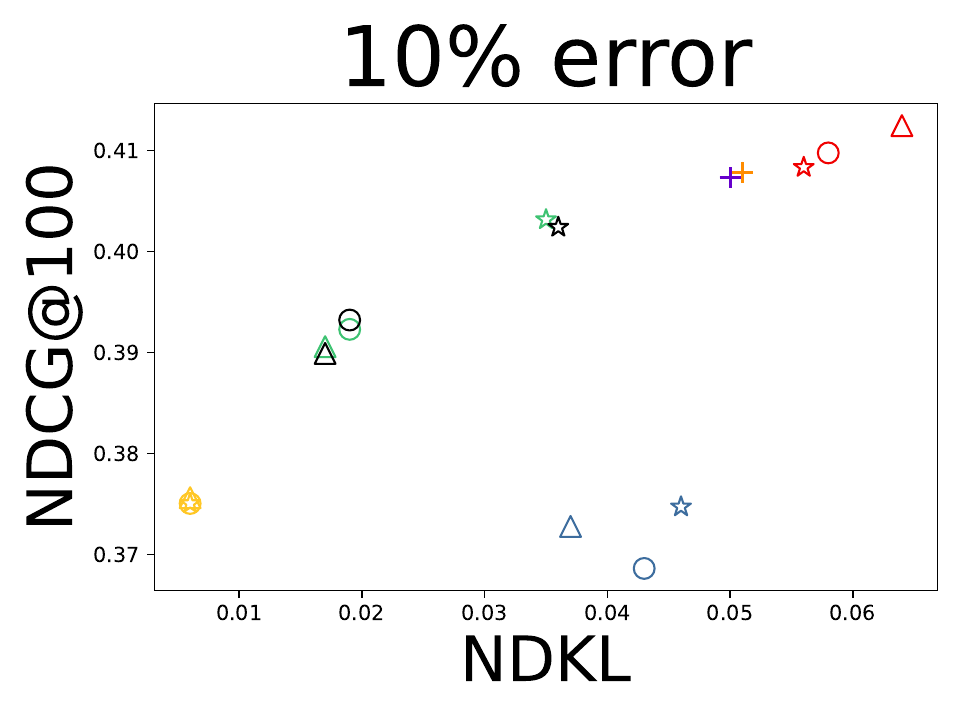}
    \end{subfigure}
    \begin{subfigure}{0.2\textwidth}    \includegraphics[width=\linewidth]{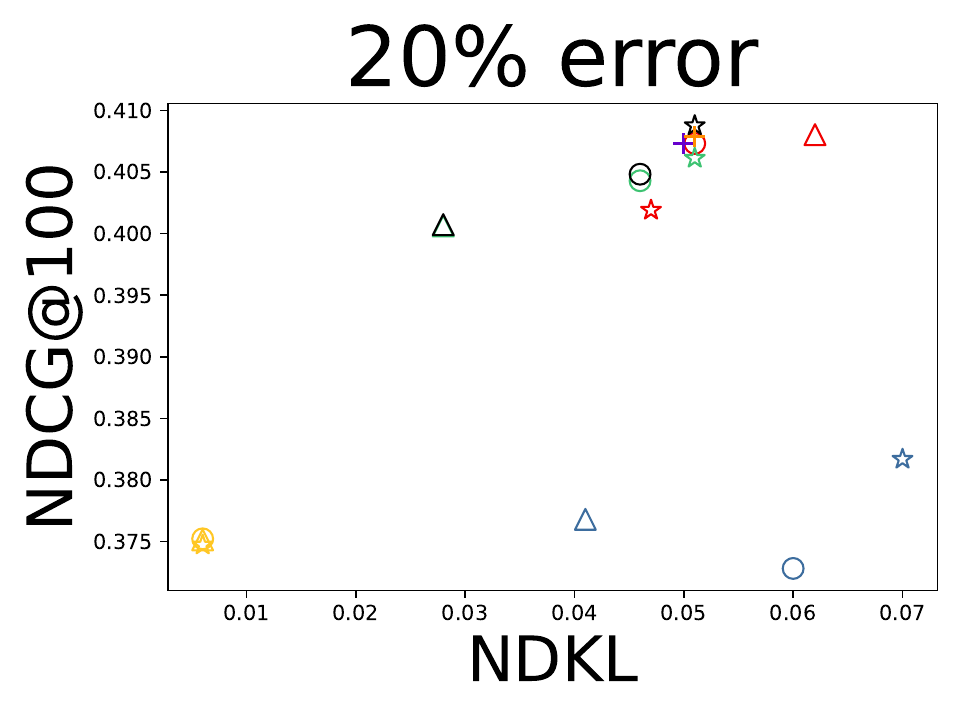}
    \end{subfigure}
    \begin{subfigure}{0.2\textwidth}    \includegraphics[width=\linewidth]{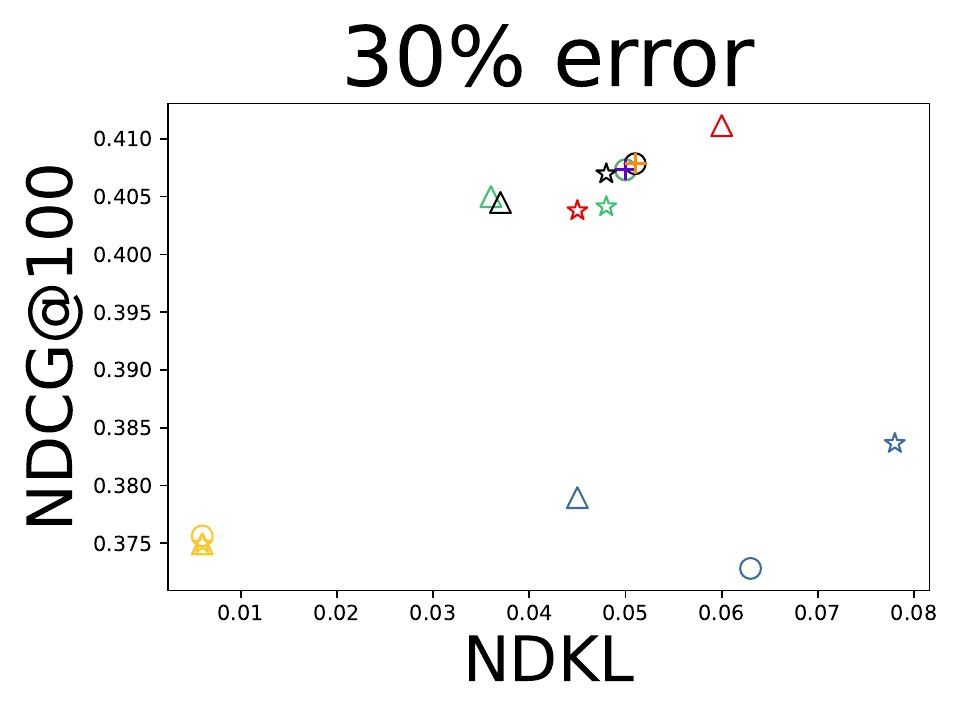}
    \end{subfigure}
    \begin{subfigure}{0.2\textwidth}    \includegraphics[width=\linewidth]{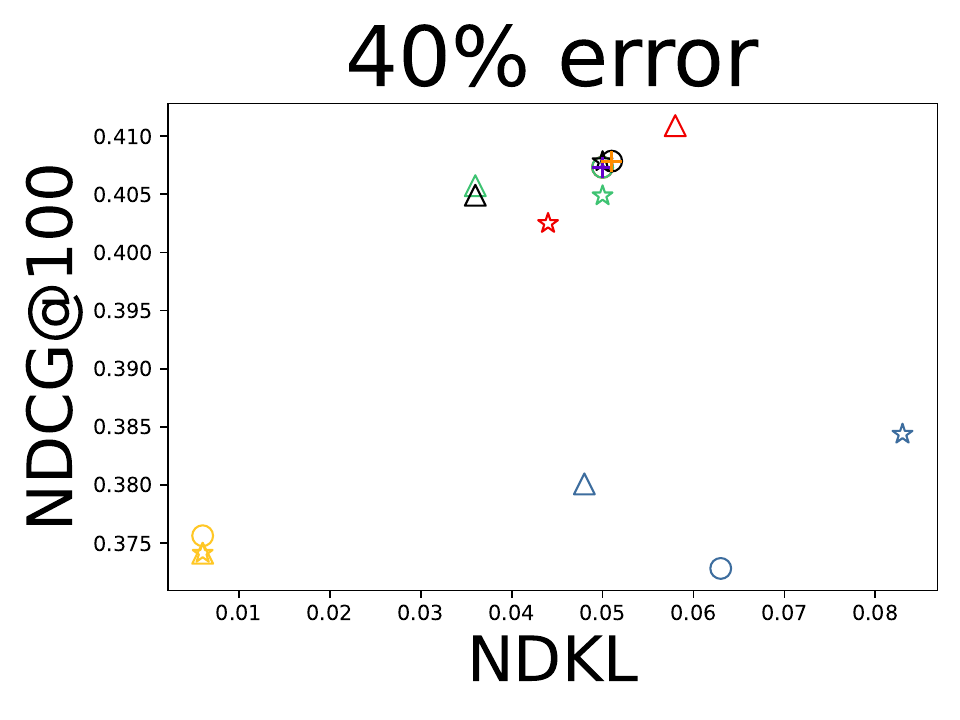}
    \end{subfigure}
    \begin{subfigure}{0.2\textwidth}    \includegraphics[width=\linewidth]{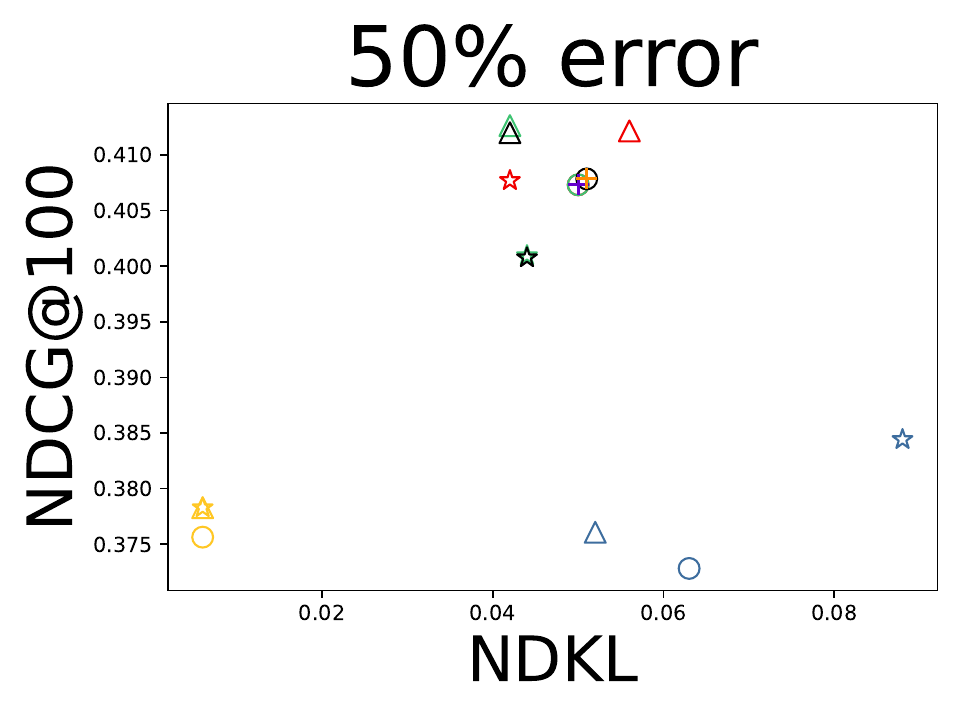}
    \end{subfigure}
    \begin{subfigure}{0.2\textwidth}    \includegraphics[width=\linewidth]{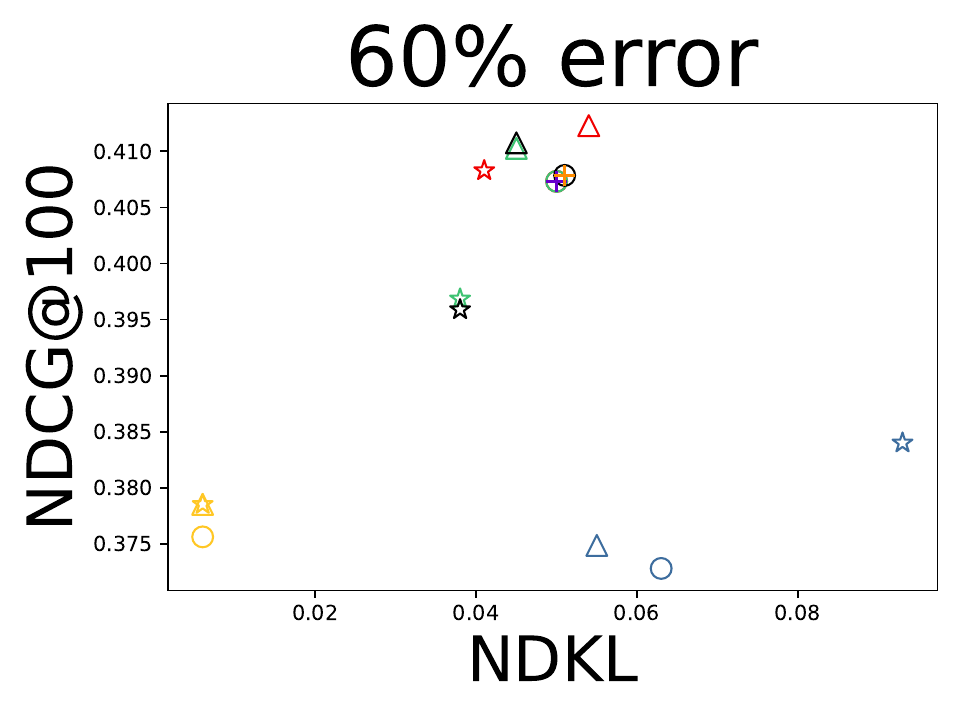}
    \end{subfigure}
    \begin{subfigure}{0.2\textwidth}    \includegraphics[width=\linewidth]{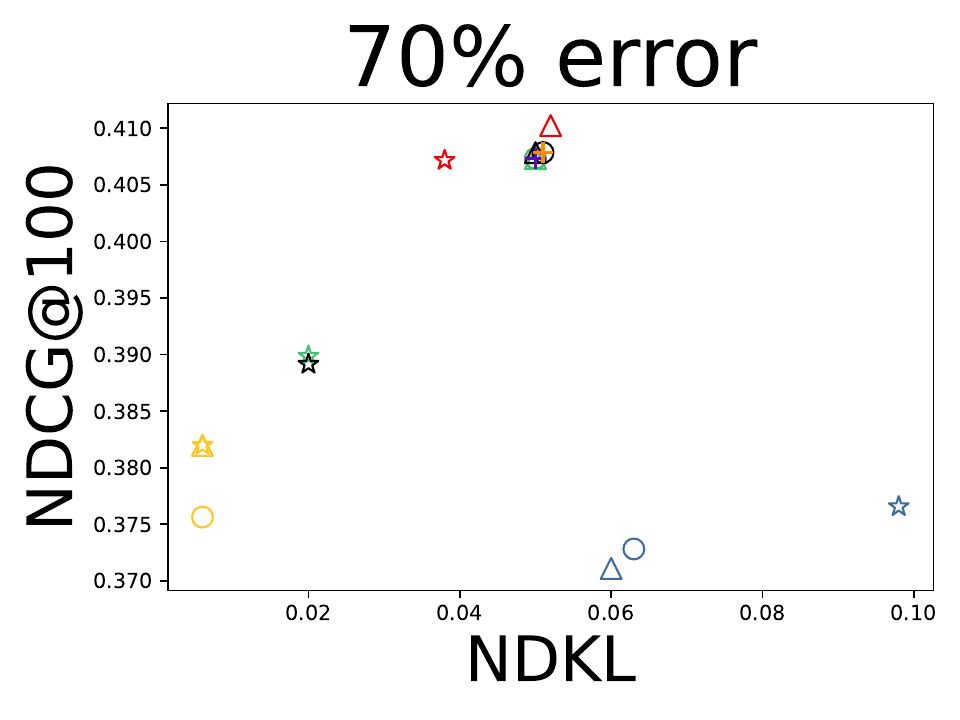}
    \end{subfigure}
    \begin{subfigure}{0.2\textwidth}    \includegraphics[width=\linewidth]{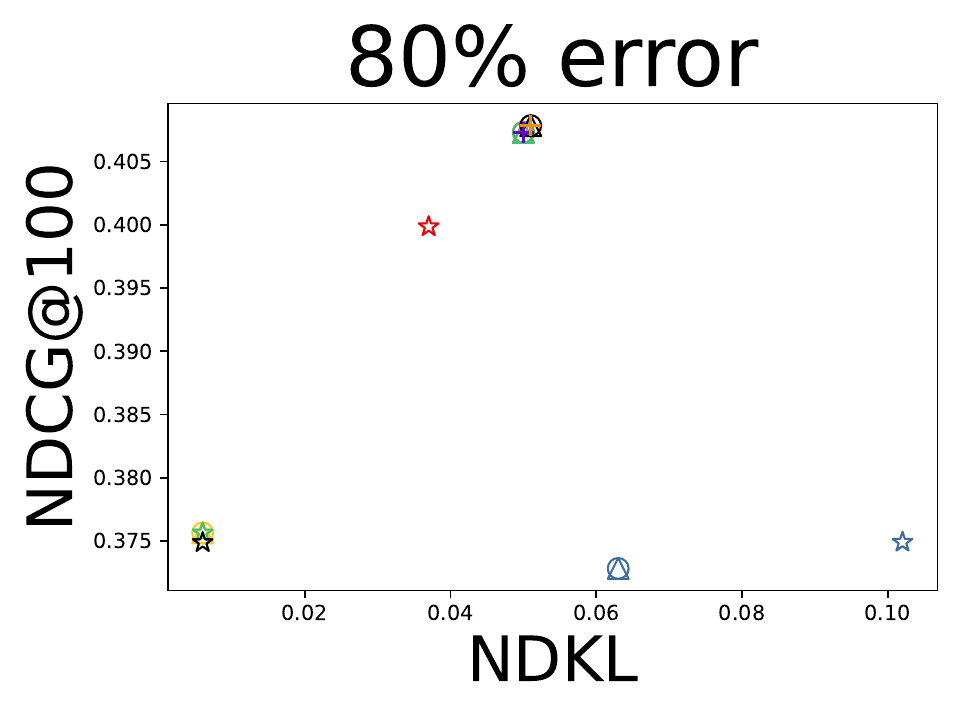}
    \end{subfigure}
    \begin{subfigure}{0.2\textwidth}    \includegraphics[width=\linewidth]{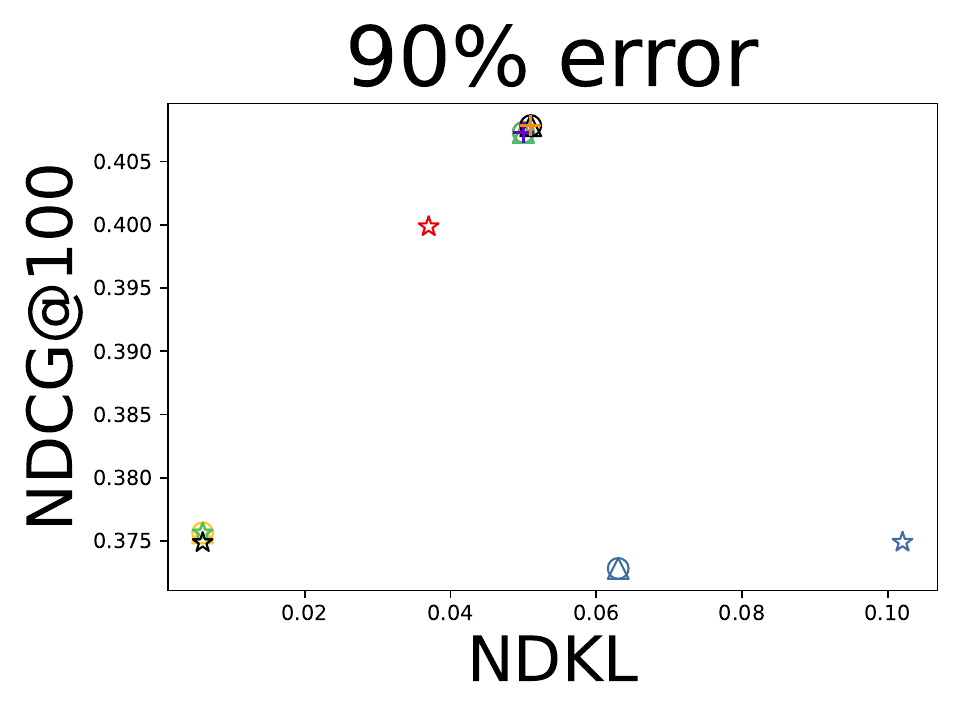}
    \end{subfigure}
    \begin{subfigure}{0.2\textwidth}    \includegraphics[width=\linewidth]{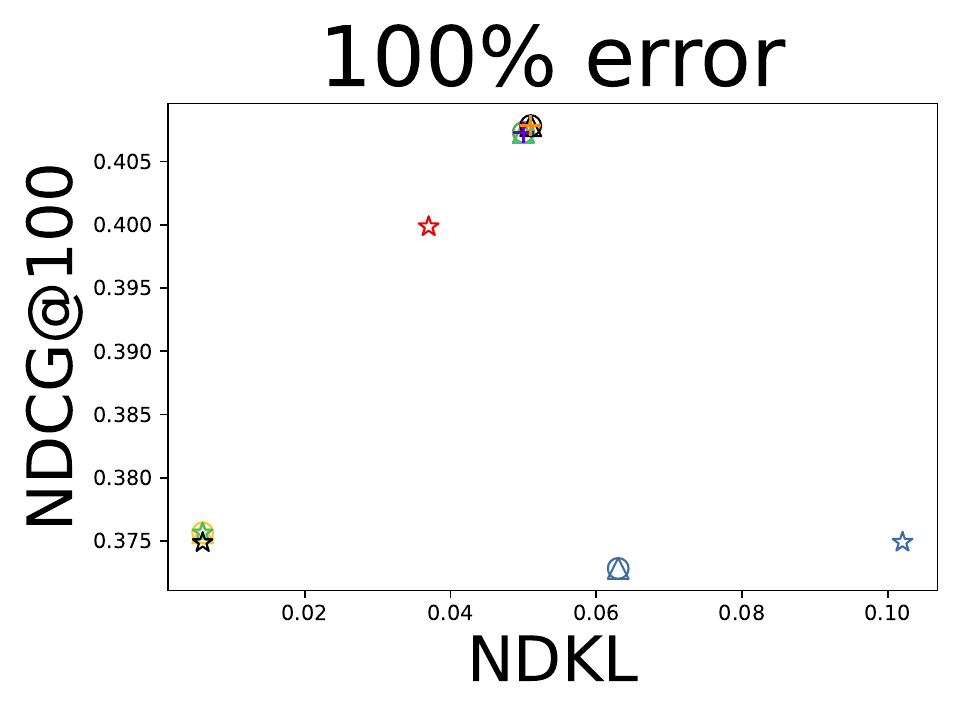}
    \end{subfigure}

    \caption{NDKL/NDCG 100 Trade-off graphs for each model for (W)NBA}
    \label{fig:tradeoff}
\end{figure*}

To get some more insights into the trade-off between NDKL as a measure of fairness and NDCG as a measure of utility, we present trade-off graphs in Figure \ref{fig:tradeoff}. The graphs show that LTR-FairRR may exhibit the least susceptibility to inference errors when considering the fairness-utility trade-off. Note that this is only with regards to the (W)NBA dataset. Also, for FairLTR, $g_{dis}\leftarrow g_{adv}$ has the potential to sustain a higher degree of fairness even under varying rates of inference errors.

\end{document}